\pdfoutput=1
\RequirePackage{fix-cm}
\documentclass[twocolumn]{svjour3}          
\smartqed  
\usepackage{graphicx}
\usepackage{multirow}
\usepackage{epsfig}
\usepackage{epstopdf}
\usepackage{amsmath}
\begin{document}

\title{Self Organizing Classifiers: First Steps in Structured Evolutionary Machine Learning}
\titlerunning{SOC: First Steps in Structured Evolutionary Machine Learning}


\author{Danilo Vasconcellos Vargas \and  
	Hirotaka Takano \and 
	Junichi Murata 
}

\institute{Danilo Vasconcellos Vargas \at
	Kyushu University, Fukuoka, Japan\\
              \email{vargas@cig.ees.kyushu-u.ac.jp}           
           \and
	Hirotaka Takano \at 
	Kyushu University, Fukuoka, Japan\\
              \email{takano@cig.ees.kyushu-u.ac.jp}           
	   \and
	Junichi Murata \at 
	Kyushu University, Fukuoka, Japan\\
              \email{murata@cig.ees.kyushu-u.ac.jp}           
}

\date{Received: date / Accepted: date}

\maketitle

\begin{abstract}
Learning classifier systems are evolutionary machine learning algorithms, flexible enough to be applied to reinforcement, supervised and unsupervised learning problems with good performance. 
Recently, self organizing classifiers were proposed which are similar to learning classifier systems but have the advantage that in its structured population no balance between niching and fitness pressure is necessary. 
However, more tests and analysis are required to verify its benefits.
Here, a variation of the first algorithm is proposed which uses a parameterless self organizing map (SOM). 
This algorithm is applied in challenging problems such as big, noisy as well as dynamically changing continuous input-action mazes (growing and compressing mazes are included) with good performance.
Moreover, a genetic operator is proposed which utilizes the topological information of the SOM's population structure, improving the results.
Thus, the first steps in structured evolutionary machine learning are shown, nonetheless, the problems faced are more difficult than the state-of-art continuous input-action multi-step ones.
\end{abstract}




\section{Introduction}
\label{intro_sec}


Learning Classifier Systems (LCS) are several algorithms inspired by evolution \cite{urbanowicz2009learning},\cite{lanzi2000roadmap}.
They can be applied to reinforcement learning problems (actually, they can solve supervised learning and unsupervised learning \cite{tamee2007towards} as well
 but we will only focus on reinforcement learning in this article). 
Different from most reinforcement learning algorithms, however, LCS algorithms do not use state-action look-up tables to predict payoff.
To solve RL problems, LCS systems use a set of individuals with condition-action-prediction rules, i.e., solving the problem with piecewise approximations \cite{lanzi2005xcs}.
In this manner, the difficulties that arrive from complex problems, where a large number of states and/or actions are required, can be avoided.

However, the dynamic niches\footnote{In this article, the term niche follows the Hutchinsonian niche definition \cite{Hutchinson1957}. Hutchinsonian niche is an $n$-dimensional hyper-volume composed of environmental features. Moreover, niching is basically clustering, i.e., the objective is to create niches (clusters) which are more similar in some sense.}
of solutions present in LCS allows over-generalized solutions with higher fitness to compete and win against specialized ones in low-fitness niches (niches where good solutions receive low payoff when compared to other niches), even though the specialized solutions would have a better performance. 
One way of solving this problem is to separate a fitness defined on a niche from fitnesses defined on other niches (i.e., having a good fitness on other niches would not influence the present niche). 
This is exactly the niched fitness concept which was introduced and used by Self Organizing Classifiers (SOC) (niched fitness is explained in Section~\ref{novel_concepts}). 
Notice that niched fitness requires well defined niches where the fitness of individuals can be measured and compared locally inside each niche.
This concept is difficult to insert into current LCSs.

SOC are a recently proposed family of evolutionary machine learning methods with a structured population (to the knowledge of the authors the first evolutionary machine learning algorithm to use a structured population).
The main objective of SOC's construction was to overcome over-generalization problems (over-generalized solutions and related problems) in LCS \cite{vargas2013self}.
In SOC systems, no balance between specialization and generalization are needed, since the niched fitness concept is used. 
Actually, the niched fitness concept indirectly requires a separation between the niching pressure from the fitness pressure, in other words, by requiring well defined niches, both objectives (i.e. the objective of creating good niches and the objective to reach good solutions) need to be clearly stated.
SOC uses evolution only for finding good solutions (fitness pressure), letting the SOM face the clustering objective in parallel (niching pressure).
Indeed, it is similar to coevolutionary approaches.
The fitness for each niche can be strength based\footnote{Strength based fitness that is directly proportional to the payoff. They came into disuse in the LCS literature because of consequential over-generalization problems.} 
because the niched fitness already solved the problems with over-generalized classifiers.

Regarding the relationship between SOC and LCS.
Although, SOC and LCS possess many similarities, it may not be correct to classify SOC as an LCS.
There are many crucial differences that make SOC difficult to match with other LCS algorithms, see Table~\ref{soc_lcs}.

\begin{table*}
\centering
\caption{Difference Between Systems}
\begin{tabular}{|c|c|l|c|} \hline
	Algorithm Type  & Model & Fitness \\ \hline
	     LCS & set of condition-action-prediction rules & Accuracy based \\ \hline
      SOC & dynamic state-table of prediction rules & Strength based \\ \hline
 Standard RL & static state-action look-up tables & Strength based \\ \hline
\end{tabular}
\label{soc_lcs}
\end{table*}
	

This article extends \cite{vargas2013self} to include a more robust method without modifying its simplicity.
A parameterless SOM replaces the SOM algorithm conferring better adaptation properties and less parameters. 
Previously, SOC were applied to some complex multi-step RL problems with optimum or near optimum results even using small population sizes.
This article shows results on four new challenging problems:
\begin{itemize}
	\item Big mazes - Mazes with as much as four times the area of previous mazes.
	\item Noisy mazes - Mazes with noise.
	\item Changing mazes - Mazes which have their structure modified over a series of trials.
	\item Growing and compressing mazes - Mazes which increase or decrease in size and structure over trials.	
\end{itemize}
Additionally, a new genetic operator is proposed which utilizes the topological information of the SOM population. 
Its use is shown to improve the results.


\section{Learning Classifier Systems in Multi-step and/or Continuous\\ Problems}

LCS have been developed for a while, forming a wide and diverse literature. 
LCS are evolutionary based systems capable of solving problems by evolving a set of agents with condition-action-prediction rules which cooperate or compete with themselves.
Here we will briefly review LCS applied to multi-step and/or continuous problems.
For a detailed review of the literature, please refer to \cite{urbanowicz2009learning}\cite{lanzi2000roadmap}.

In problems with continuous actions, LCS was applied to many problems.
To begin with, XCSF has been applied to function approximation \cite{wilson2002classifiers},\cite{butz2008function},\cite{tran2007xcsf}.
Other works in function approximation include the LCS with fuzzy logic  \cite{valenzuela1991fuzzy},\cite{bull2002accuracy},\cite{casillas2007fuzzy}, neural-based LCS algorithms \cite{bull2002using},\cite{bull2002accuracy} and genetic programming-based \cite{iqbal2012xcsr}.
The success of LCS also span the control of robotic arms  \cite{stalph2012learning,butz2008context} and navigation problems \cite{bonarini2000fuzzy,howard2009towards}.

However, applications to multi-step problems with continuous actions restrict to the mobile robot in a corridor \cite{bonarini2000fuzzy} and the empty room with noise \cite{howard2009towards}.
Complex multi-step problems were solved only for \textbf{discrete outputs} \cite{lanzi2005xcs}.

\section{The Parameterless Self Organizing Map}

\begin{figure}
\centering
\includegraphics[height=0.9in]{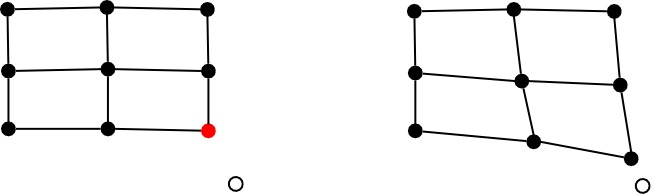}
\caption{Illustration of the SOM dynamics. The left part of the figure shows a two dimensional input (white dot) being presented to the SOM's grid (black dots connected by lines). In this scenario, the winning cell (closest to the input) is shown in red. Afterwards, the SOM's grid is updated and the resulting grid is shown on the right.}
\label{som_illustration}
\end{figure}

The commonly used SOM is an algorithm capable of producing a projection of the input usually into two dimensional space.
One of the most important advantages of the method is the preservation of the topological relationship of the input in the constructed map.
In SOM, for every input $x$ the grid of weights $w_i$ compete for it (the closest weight wins the competition, i.e., the weight having minimum $||x-w_i||$).
After the winning cell is decided, all cells in the grid are updated by the following equation:
\begin{align}
	&\Delta w_i(t)= \epsilon(t)h_{i,c}\{x(t)-w_i(t)\}\\
	&w_i(t+1)= w_i(t)+\Delta w_i(t).
\end{align}
where $\Delta w_i(t)$ is the weight update in the current iteration $t$, $\epsilon$ is the learning rate and $h_{i,c}$ is the neighborhood function. $x$ and $w_i$ are respectively the input array and the weight of a given cell $i$ when the winning cell index is $c$.
The learning rate $\epsilon$ is a monotonically decreasing function with respect to the number of times the grid update was realized and neighborhood function is usually an exponentially decreasing distance based function.  
For example:
\begin{align}
&\epsilon = 0.1(0.999999)^{t}\\
&h_{i,c} = e^{-dist(i,c)^2},
\end{align}
where $dist$ is some distance metric applied on the grid, defining the topological relationship of the grid's cells.
Figure~\ref{som_illustration} illustrates one iteration of the algorithm, showing how the grid adapts to a given two dimensional input.

A parameterless SOM is a SOM where the learning rate function is not necessary \cite{berglund2006parameterless}.
Instead, it is automatically defined by how good the SOM fits the input (in fact, this value is strongly related to the input's novelty \cite{ICDL12-hester},\cite{reehuis2013novelty}).
In the usual SOM, by making learning rate monotonically decreasing with the number of iterations, the more the SOM is used the less it could learn/adapt to new information.
The parameterless SOM sets the learning rate according to the error of the input (not a monotonic decreasing function), therefore it is able to increase the learning rate when an rare or unexpected input arrives.
Thus, the new parameterless SOM does not only possess fewer parameters, but also the ability to always adapt to changes in the environment.

Let the learning rate $\epsilon$ be defined as:
\begin{align}
	&r(0)=||x(0)-w_c(0)||\\
	&r(t)= max(||x(t)-w_c(t)||,r(t-1))\\
	&\epsilon(t)= \frac{||x(t)-w_c(t)||}{r(t)},
\end{align}
where $w_c$ is the SOM winning cell's weight array.
The weight update $\Delta w_i(t)$ is similar to SOM's weight update, changing only in relation to the new learning rate $\epsilon$ and the modified neighborhood function $h_{i,c}$.
Considering $dist(i,c)$ the distance between cell $i$ and the winner cell $c$, we have:
\begin{align}
	&\Theta(\epsilon(t)) = \epsilon(t)\theta_{max}, \Theta(\epsilon(t))>\theta_{min}\\
	&h_{i,c}=e^{\frac{-dist(i,c)^2}{\Theta(\epsilon(t))^2}}
\end{align}
$\theta_{max}$ and $\theta_{min}$ are respectively the maximum and minimum of $\Theta(\epsilon(t))$.
In this article, $\theta_{max}$ equals to the SOM's area (width multiplied by the height of the grid) and $\theta_{min}=1$ are used.

\section{Structured Evolutionary Algorithms}
\label{struct_ea}

Structured evolutionary algorithms does not possess a panmictic population.
Instead they organize the individuals into a structured population \cite{tomassini2005spatially,alba2002parallelism}.
As commonly considered in the literature, algorithms which has some sort of implicit structure will not be considered structured.
In fact, to avoid this type of confusion the name of parallel evolutionary algorithms are sometimes used in the literature.
The need for distinction derives from the fact that algorithms with implicit structure lose many of the benefits of ones with explicit structure.

Two types of structured EAs will be given as examples which are somewhat related to the structure of the proposed method.

\begin{figure}
\centering
\includegraphics[height=1in]{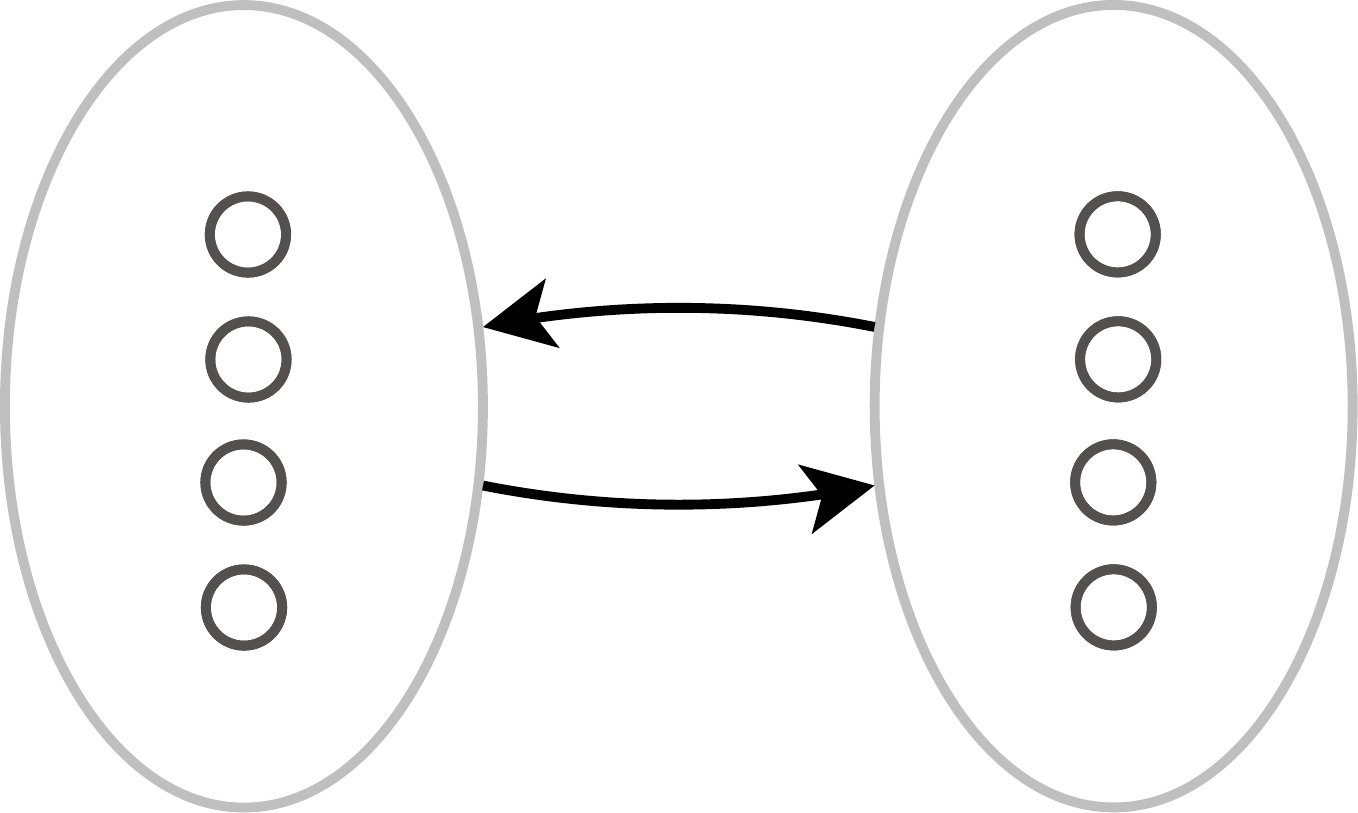}
\caption{Island model structure. Arrows indicate the infrequent immigration procedure, the circles are the individuals and the oval shapes are the subpopulations.}
\label{island}
\end{figure}

The first type is island models (also called distributed genetic algorithms) \cite{belding1995distributed}.
Figure~\ref{island} shows its structure. 
Basically, the population is divided into a number of subpopulations (``islands") with few genetic information exchanged between them.

\begin{figure}
\centering
\includegraphics[height=1in]{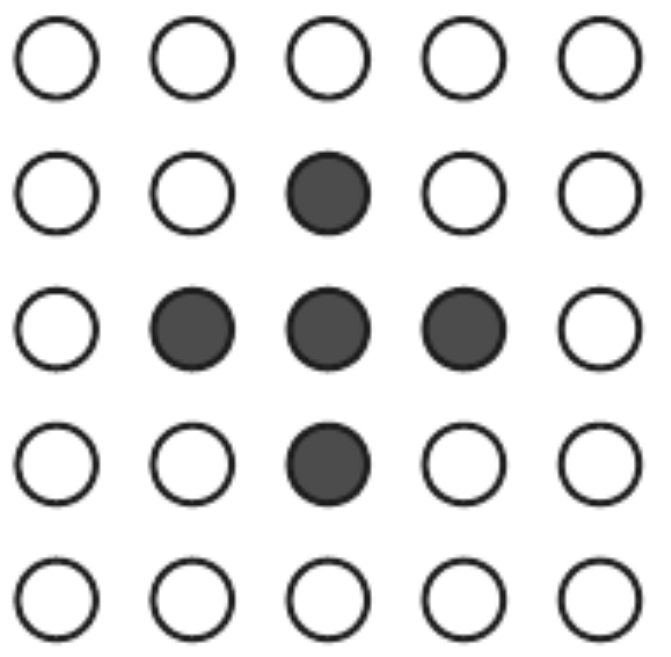}
\caption{Cellular algorithm structure. Shaded area indicates an example of neighborhood for the central individual.}
\label{cellular}
\end{figure}

The second type, cellular algorithms are structured evolutionary algorithms where individuals are usually positioned in a vertex of a lattice graph (Figure~\ref{cellular} shows a common cellular structure). 
They interact solely with adjacent individuals defined by a neighborhood function \cite{manderick1989,alba2000cellular}.

\section{Evolutionary Machine Learning - Structured and Unstructured}
\label{structured}

\begin{figure}
\centering
\includegraphics[height=1.6in]{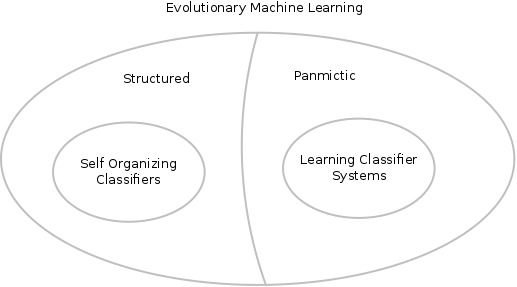}
\caption{Structured and unstructured division.}
\label{struct_field}
\end{figure}

Methods in evolutionary machine learning can also be classified into structured and unstructured\footnote{Structured and parallel as well as unstructured and panmictic terms when referring to algorithms will be used indistinctly.}.
As explained in Section~\ref{struct_ea}, structured algorithms arrange their population in some sort of structure whereas unstructured ones possess a single population set.
That is, in the same way that structured evolutionary algorithms differ from unstructured panmictic ones, learning classifier systems can be classified, in this context, as a type of unstructured evolutionary algorithms and self organizing classifiers can be seen as a structured evolutionary algorithm.
Figure~\ref{struct_field} shows a diagram illustrating this aspect.
Note that implicit structured algorithms such the niched genetic algorithm can not be considered structured following the definition in Section~\ref{struct_ea}.


There is a motivation behind structured evolutionary machine learning.
Indeed, some advantages of structured against unstructured algorithms should follow the ones from the optimization field.
To cite a few:
\begin{itemize}
	\item Diversity - By restricting the interrelation of individuals, structured populations can preserve different subpopulations, i.e., avoid competition;	
	\item Time Performance - Structured populations allow easier parallelization and therefore less running time when running in the appropriate system.
\end{itemize}

\section{Novel Concepts}
\label{novel_concepts}

\begin{figure}
\centering
\includegraphics[height=1.5in]{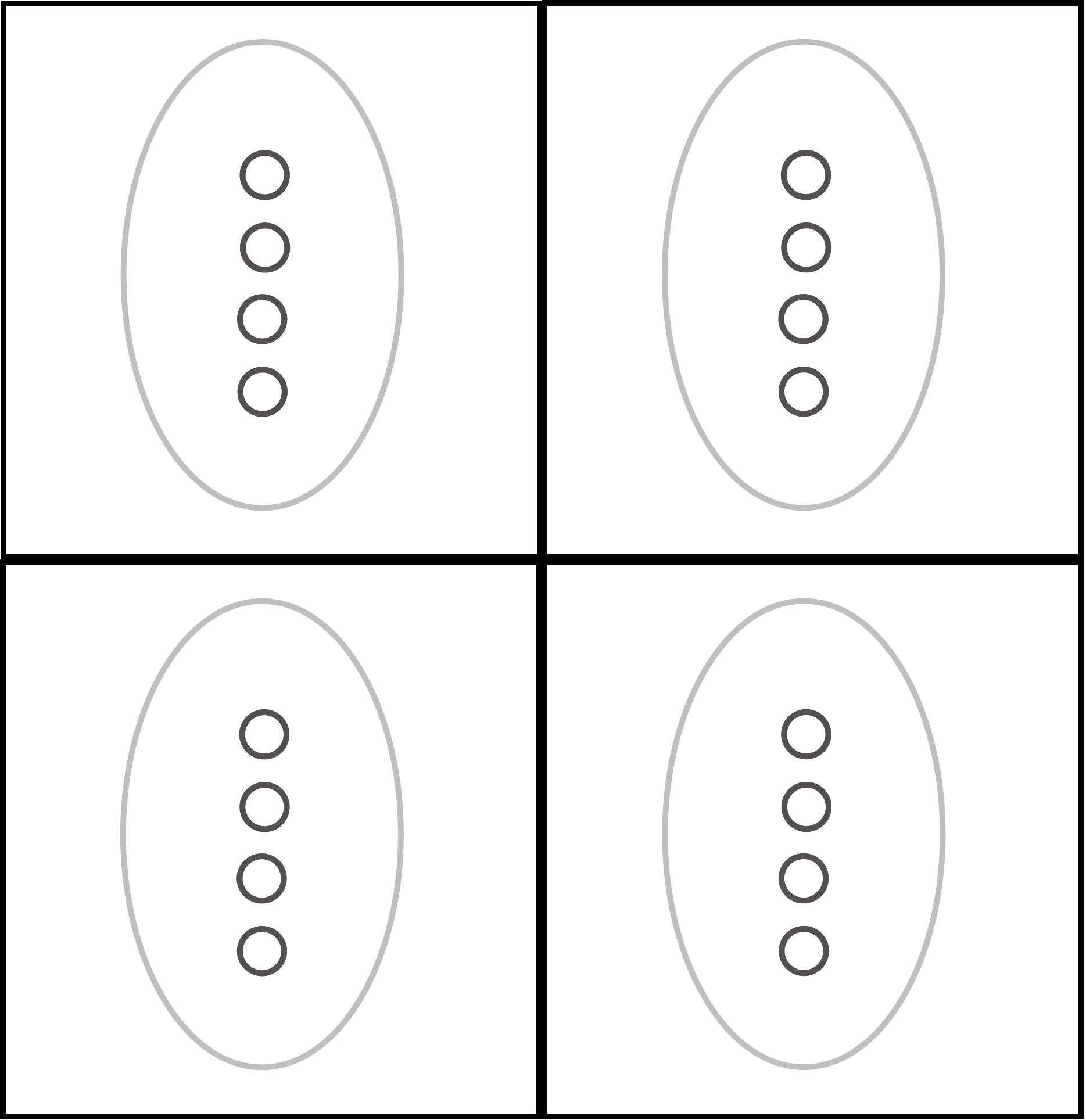}
\caption{SOM population structure. A self organizing map grid with subpopulations inside each cell.}
\label{som_pop}
\end{figure}

Two relatively novel concepts are used in the article, both concepts were introduced in \cite{vargas2013self}. They are respectively:
\begin{itemize}
	\item Niched fitness - Niched fitness is a fitness defined only in a given niche (place or circumstance). Outside of this niche, the fitness is undefined and therefore nonexistent. The objective here is to evaluate individuals by their performance in the given niche independent of how they behave in other niches, solving over-generalization problems. In fact, the niched fitness concept expose the similarity between niching and multiobjectivization \cite{mouret2011novelty}. 
	\item SOM population -  SOM population is a 2D cell grid with each cell having a subpopulation. The grid behave as a SOM, self-organizing itself to the input and allowing only the subpopulation inside the winning cell to interact with the input. In other words, SOM population is a mixture of both island models and cellular algorithms with a self-organizing structure (See Figure~\ref{som_pop}).
\end{itemize}

\section{Self Organizing Classifiers}
\label{soc}

\begin{figure*}
\centering
\includegraphics[height=3in]{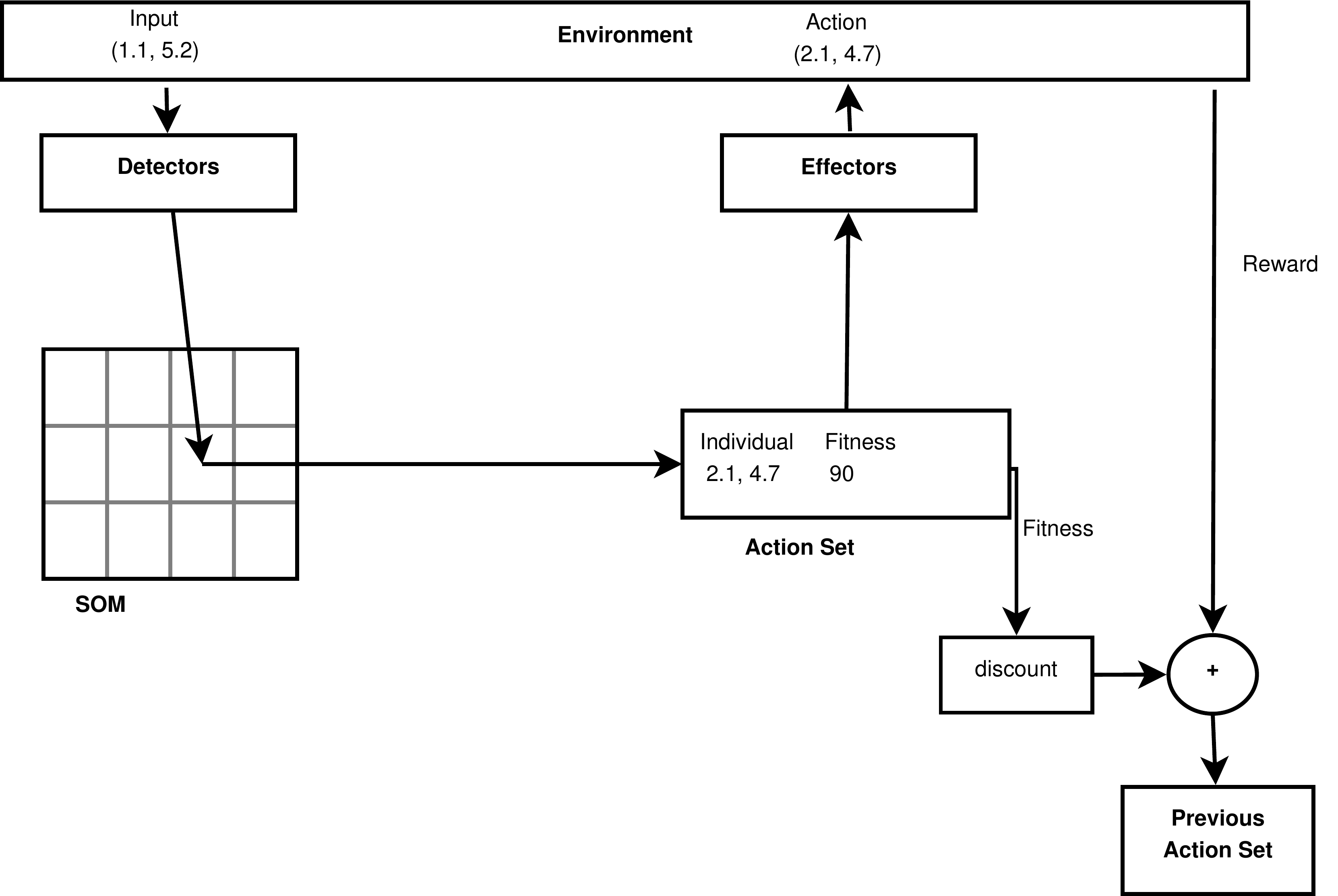}
\caption{Self Organizing Classifier Schematic}
\label{schematic}
\end{figure*}

SOC are a series of algorithms based on the SOM subpopulation, sharing similar model and dynamics.
The schematic of SOC is described in Figure~\ref{schematic} (the schematic style is the same as the one previously used to describe ZCS, XCS and many other papers of the LCS literature \cite{wilson1994zcs}).

A Q-learning based reinforcement scheme with niched fitness is used by SOC.
The fitness update of each individual is done using the Widrow-Hoff rule \cite{Widrow1960Adaptive}:
\begin{equation}
F = F + \eta(\hat{F} - F),
\end{equation}
where $\eta$ is the learning rate, $F$ is the current fitness and $\hat{F}$ is a new fitness estimate.
Given an arbitrary classifier $c$ activated at its SOM's cell $cell$.
The fitness estimate of the pair ($cell$,$c$) which were activated at time $t-1$ is given by the following equation:
\begin{equation}
	\hat{F}(c,cell)_{t-1} = R_{t-1} + \gamma \underset{c' \in cell'}{\operatorname{max}} \{F(c',cell')\},
\end{equation}
where $R$ is the reward received, $\gamma$ is the discount-factor,
$\underset{c \in cell}{\operatorname{max}} \{F(t)\}$ is the maximum fitness inside the activated cell $cell'$ at the current cycle $t$ and $c'$ is a classifier which has the current highest fitness in $cell'$.

Similar to learning classifier systems, to decrease computation resources, the structure of the SOM population is implemented as a single array of classifiers with a given numerosity indexed by the SOM population structure.
In this manner, the numerosity is defined by the number of indexes a given individual possesses.

\section{Why SOM?}

Before going into the further details of the algorithm, we wish to clarify one question that might surge.
Among the vast amount of algorithms that could cluster inputs, why was SOM chosen?

The answer is that we chose SOM not for what it does, but for what it gives as information.
In this context, SOM's interesting capabilities are as follows:
\begin{itemize}
	\item Topological Preserving Projection - high-dimensional inputs are projected in a two-dimensional topological preserving map.
	\item Novelty Measure - The error of the SOM's cell given the input is an approximation to the uniqueness of the input itself, which is a measure of novelty \cite{ICDL12-hester}, \cite{reehuis2013novelty}.
In fact, this measure is used to affect the update of the other cell weights in the parameterless SOM.
\end{itemize}		

Evolutionary algorithms can use these pieces of information to better adapt to the problem at hand.
Moreover, other procedures can be run on these information to improve the system as a whole.
For example, genetic operators may exploit the SOM's structure by reproducing individuals with similar individuals in subpopulations adjacent to them.
Observe that adjacent subpopulations are necessarily similar in input and therefore may share similar or even equal solutions.

\section{Simplest Self Organizing Classifiers}

Simplest Self Organizing Classifiers are implementations of the class of Self Organizing Classifiers (see Section~\ref{soc}).
Figure~\ref{schematic} shows how the cycle is executed and Table~\ref{ssoc_table} describes the execution cycle in details.

\begin{table}[h]
 \centering
\caption{Simplest Self Organizing Classifiers' Cycle} 
\begin{tabular}{p{8cm}}
\hline
\begin{enumerate}
\item An input is received by the system
\item The SOM population is activated on the input (a given individual will be returned to act)
\begin{enumerate}
\item The cell's weight array which is closest to the received input wins the competition 
\item Inside the winning cell, a random individual is chosen either from the novel group or from the best group (depending if it is an exploration or exploitation cycle)
\item The cells' weight array of the SOM population is updated by the SOM algorithm
\item The chosen individual is returned
\end{enumerate}
\item The chosen individual acts on the system (in the case of SSOC, the individual's chromosome is the action itself).
\item The previous acted individual on the past activated cell has its fitness updated. The equation used to update is written in Section~\ref{soc} (notice that the fitness is updated only for the given cell even if the individual is present on other cells, see the niched fitness concept on Section~\ref{novel_concepts}).
\item Check if the EA should be called. If positive, execute the EA (see Section~\ref{evol_sec} for the complete description of the EA).
\end{enumerate} \\
\hline
\end{tabular}
\label{ssoc_table}
\end{table}

For the sake of comprehensibility, we give here the name of Simplest Self Organizing Classifier (SSOC) to the first very simple algorithm developed in \cite{vargas2013self} and the name of "Simplest Self Organizing Classifier 2" (SSOC2) to the one described in this article.
The method proposed here replaces the SOM by a parameterless SOM, reducing two of the parameters since the learning rate function is not necessary \cite{berglund2006parameterless}.
There are not any other differences, therefore the definition here applies to both SSOC and SSOC2.

The classifiers used code directly the action by an array of real numbers, i.e., the model is an array of real numbers which is mapped directly to the output.

Moreover, the SOM population has particularly a subpopulation inside each cell which is also divided into two groups: one of best individuals and the other of novel individuals. 
Best and novel individuals have a fixed size of $\beta$ and $\nu$ respectively.
Considering evolutionary algorithm's cycle is an algorithm cycle when the evolutionary algorithm (EA) is called, the following rules take place:
\begin{itemize}
\item Best individuals are the best fitted individuals inside the subpopulation in the last EA's cycle.
\item Novel individuals are renewed every EA's cycle (the detailed process is described in the next Section). 
\end{itemize}

The SOM population begins without any classifiers. 
Classifiers are created when the respective cell wins the competition inside the SOM.
In one hand, novel individuals are created as random classifiers.
On the other hand, best individuals, when possible, are set equal to another cell's best individuals from the neighborhood\footnote{Neighborhood is defined as the cells within a Chebyshev distance of less or equal to four} which maximize $\frac{experience}{chebyshevDistance^2}$ (experience is defined on Section~\ref{evol_sec}).
If not possible, best individuals are initialized in the same way as the novel individuals.
Observe that a given cell's experience are set to zero every time the evolution happens in it.
This may seem counter-intuitive, but it alleviates a problem with hyper active cells which often possess a high error (see the cell's error for the growth of SOM in \cite{alahakoon2000dynamic}, \cite{dittenbach2000growing}).
A random selection of a close cell should output the similar results.

Cycles of exploration and exploitation are alternated (a cycle of exploration is followed by an exploitation cycle and so on).
Within the SOM's winning cell in a giving exploration or exploitation cycle a random individual from respectively the novel or best individuals are chosen to act.

\subsection{Evolution}
\label{evol_sec}

For every cycle that a cell's individual acts, this cell has its experience counter increased.
The evolutionary algorithm is called locally on each cell when the cell's experience is greater than $\iota S$. 
$S$ is the number of subpopulation individuals (novel plus best individuals) present on each cell and parameter $\iota$ defines an experience per individual, above which they should have an accurate fitness evaluation.
After the evolution has been applied, the experience of the cell is set to zero.

By applying the evolutionary algorithm locally, SSOC2 (the same is valid for SSOC) respects the niched fitness concept.
Its procedure consists of sorting the individuals of the given cell according to their fitness.
On one hand, the current best $\beta$ individuals substitute the previous best individuals and the remaining individuals are discarded (the index is removed and the individual numerosity decrease, if the numerosity reaches $0$ it is deleted).
On the other hand, novel $\nu$ individuals are created using either:
\begin{enumerate}
	\item Indexing - A copy from (index to) a randomly selected individual of the entire population;
	\item Reproduction - Created by a genetic operator.
\end{enumerate}		
The two procedures above have equal probabilities of happening.

The differential evolution operator is used in both SSOC and SSOC2.
Motivated by its robustness and overall good results \cite{storn1997differential,vesterstrom2004comparative,iorio2005solving}, even when compared against complex optimization algorithms (e.g., Estimation of Distribution Algorithms) \cite{garcia2009study}.
In this paper, the differential evolution's mutant vector is created by randomly choosing three vectors from the SOM's entire population of individuals (notice that individuals with numerosity bigger than one are counted as one).

\section{Experiments}

The definitions done in Section~\ref{def} are important specially when discussing adaptation of the system in question.

\subsection{Definitions}
\label{def}

\newtheorem{defi}{{Definition}}
\begin{defi}
	Adaptation - Capability of a system to modify its behavior in order to better fit to changes in the environment.
\end{defi}
\begin{defi}
	Adaptation's Time - The time necessary for a system to adapt to changes in the environment.
\end{defi}

A system with an adaptation's time of zero can be said to possess a generalized model where modifications are unnecessary to fit both past and current environments. 
In this sense, generality and adaptation are closely related terms.
Although, it depends on the details about the system, e.g., the type of model used.

\subsection{Environments}

\begin{figure}
\centering
\includegraphics[height=1in]{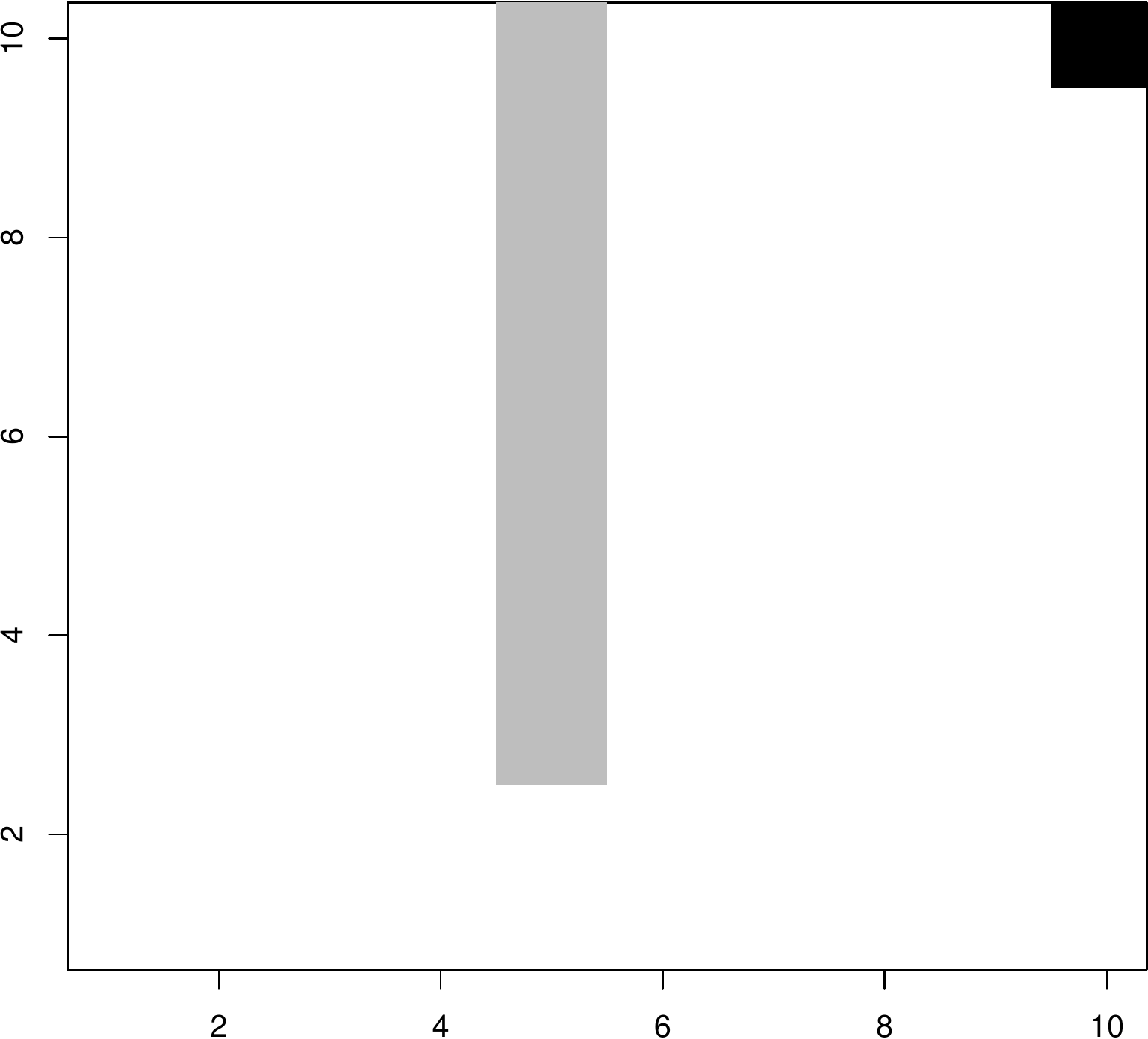}
\includegraphics[height=1in]{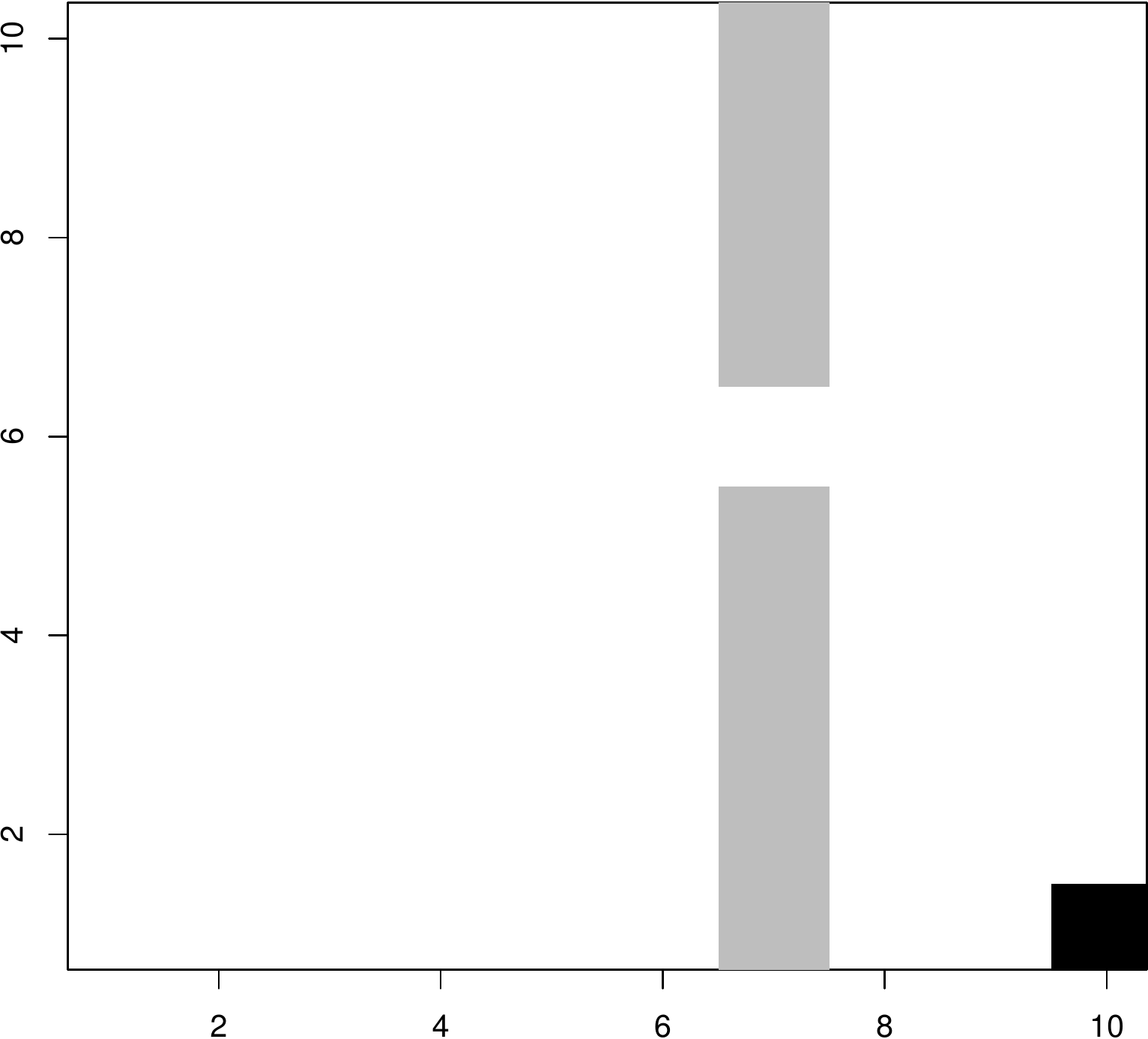}
\caption{Maze 3 - Dynamic maze problem with constant dimensions 10x10. The maze change from the left to the right state and vice-versa every $10000$ trials.}
\label{mazes2}
\end{figure}

\begin{figure}
\centering
\includegraphics[width=1in, height=1in]{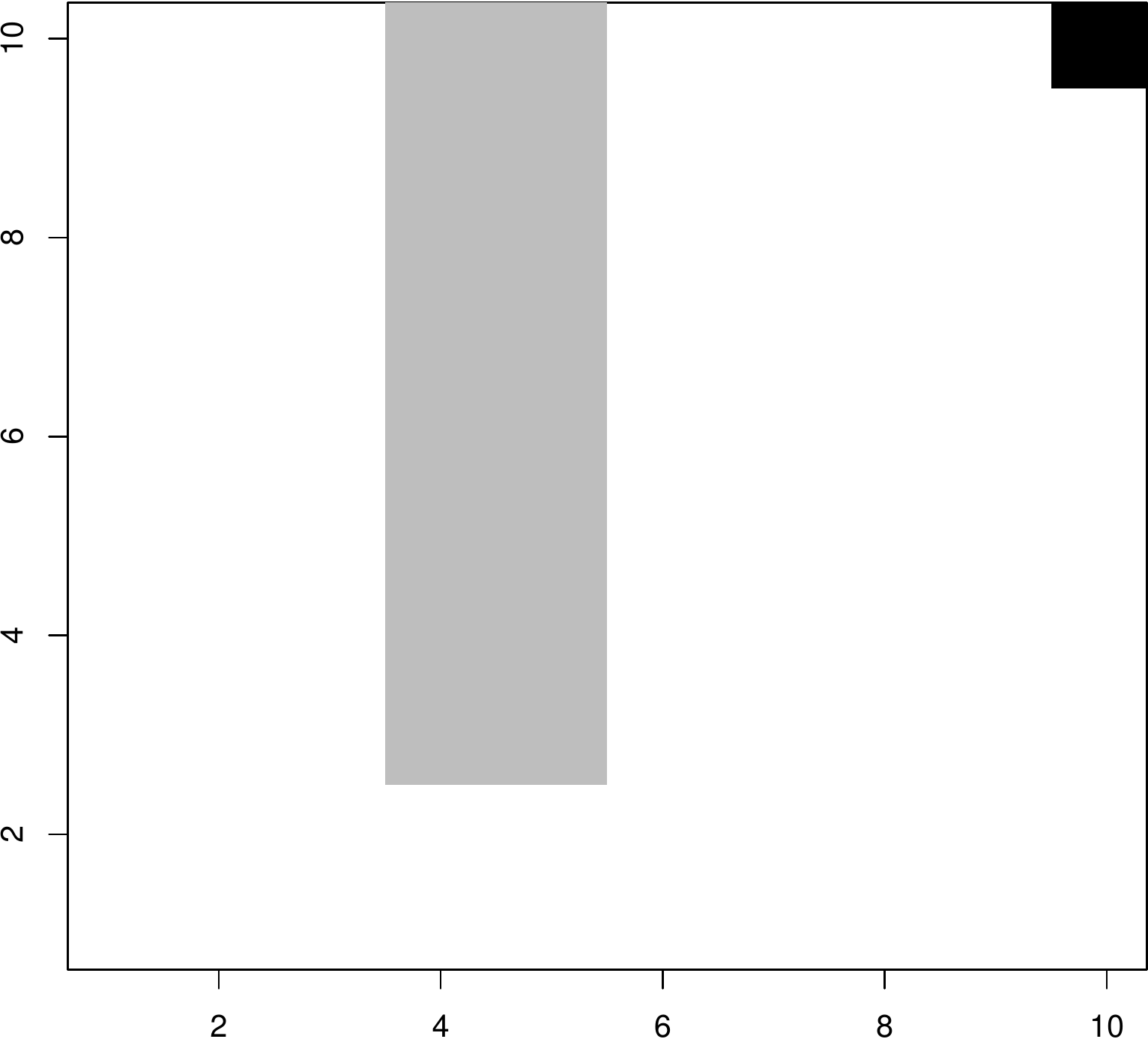}
\includegraphics[width=2in, height=1in]{maze7.pdf}
\caption{Maze 4 - Dynamic maze problem where the dimensions also change, growing from 10x10 to 10x20 state and vice-versa.}
\label{mazes3}
\end{figure}

The experiments were conducted on both mazes of Figure~\ref{mazes1} as well as on the dynamic mazes of Figures~\ref{mazes2} and~\ref{mazes3}.
Figures~\ref{mazes2} and~\ref{mazes3} show each two maze states, in each of these problems the maze is constantly changing every $10000$ trials between one state (right side) and the other state (left side).
Notice that the problem described in Figure~\ref{mazes3} shows a maze which constantly increases and decreases in size.
Agents act on all environments with continuous (x,y) translation actions. 
The variable observed by the agent is the agent's position which is also continuous. 

At every trial, the agent starts at a random position on the environment. 
Naturally, starting inside a wall is not possible.
Reaching the goal would give the agent a reward of $1000$, hitting an obstacle would return $-20$ and any other action would return $-10$.
Additionally, agents can not move more than $1.0$ in any direction.
The collision system is simply implemented (which makes it harder than a real system). 
If an agent tries to move inside a wall, the system detects the infraction, sets the agent in the previous position and returns the reward.
In other words, an agent constantly hitting the wall will not move at all.
However, an agent that hits the limits of a maze will have its final position limited by the environment.
Therefore, it is possible to move sideways when hitting the limit of the environment.


\begin{figure}
\centering
\includegraphics[width=2in, height=1in]{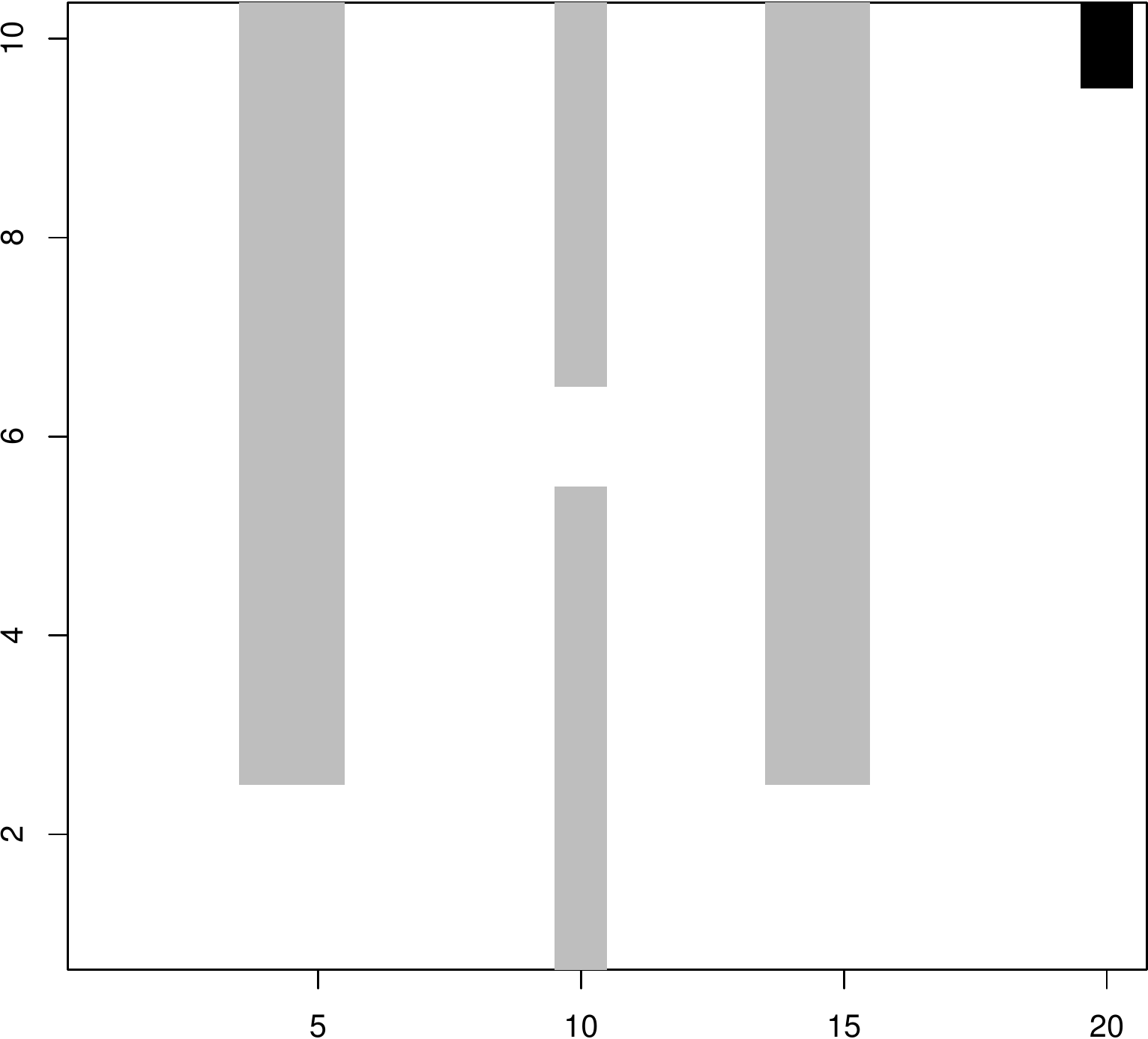}
\caption{Maze 1 - Static 10x20 maze problem.}
\label{mazes1}
\end{figure}

\begin{figure}
\centering
\includegraphics[width=2in, height=2in]{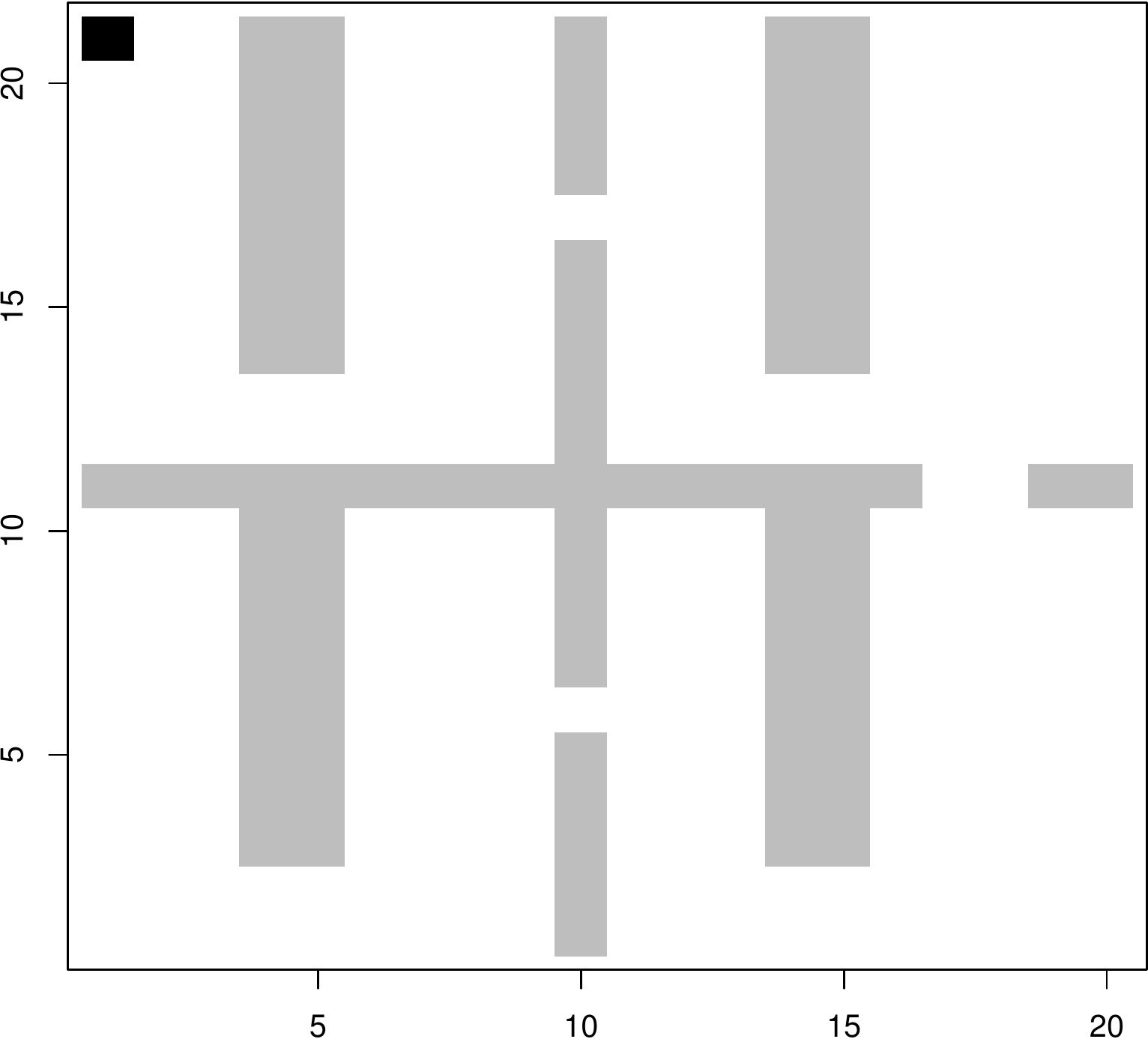}
\caption{Maze 2 - Static 21x21 maze problem.}
\label{mazes1}
\end{figure}

\subsection{Settings and Design of Experiments}
\label{exp_settings}

The parameters of the algorithm are fixed and described in Table~\ref{para}.
Here, $it$ means the SOM iteration number, $chebyshevDistance()$ is the Chebyshev distance between the current cell and the cell which won the SOM's competition and $random(a,b)$ is a function which returns a uniform random value between $a$ and $b$.
The cells of the SOM are only updated if the neighborhood function multiplied by the learning restraint surpasses the cell update threshold.

Following the design of \cite{lanzi2005xcs} the performance is computed as the average steps to reach the goal during the last $100$ trials. 
Moreover, trials can not last more than $500$ steps. 
Any trial which last more than $500$ is terminated and a new trial is started with the agent, as usual, in a random position.
Every result, when not stated otherwise, are averaged over $20$ experiments.

To give an overall idea of the resulting behavior of the agent (or its fitness) after the algorithm has learned, a sampling strategy was used.
First, we divide the maze in blocks of size $1$x$1$, then the action of the agent (or the fitness of the winning cell) is sampled $100$ times and averaged in a given $1$x$1$ block.
By repeating this process for all blocks inside the maze, we have a general idea of the behavior learned (or fitness distribution).

\begin{table*}
\centering
\caption{Parameters}
\begin{tabular}{ |c|l|l| }
	\hline
	\multirow{2}{*}{Differential Evolution} & CR & $0.2$\\
	& F & $random(0,1)$ \\ \hline
	\multirow{5}{*}{Self Organizing Map} & Matrix Size & $10\times10$ \\
	 & Weight's initial value & $random(0,1)$ \\ 
	 & Neighborhood function & $\exp(-chebyshevDistance()^2)$ \\
	 & Cell update threshold & $0.005$ \\
	\hline
	\multirow{5}{*}{Self Organizing Classifiers} & $\eta$ & $0.2$ \\ 
	 & $\beta$ & $2$ \\
	 & $\nu$ & $5$ \\
	 & $\iota$ & $20$ \\
	 & $\gamma$ & $0.90$ \\
	 & $Initial Fitness$ & $0$ \\
	\hline
\end{tabular}
\label{para}
\end{table*}

\subsection{Big Mazes}

Big mazes challenge methods to create long chains of actions.
In fact, the creation of long chains of actions may be difficult for various reasons.
For example, fitness may not spread wisely, learning may take too long (e.g., due to maze bottlenecks), instabilities may arise, etc.

\begin{figure*}
\centering
\includegraphics[width=3in, height=1.5in]{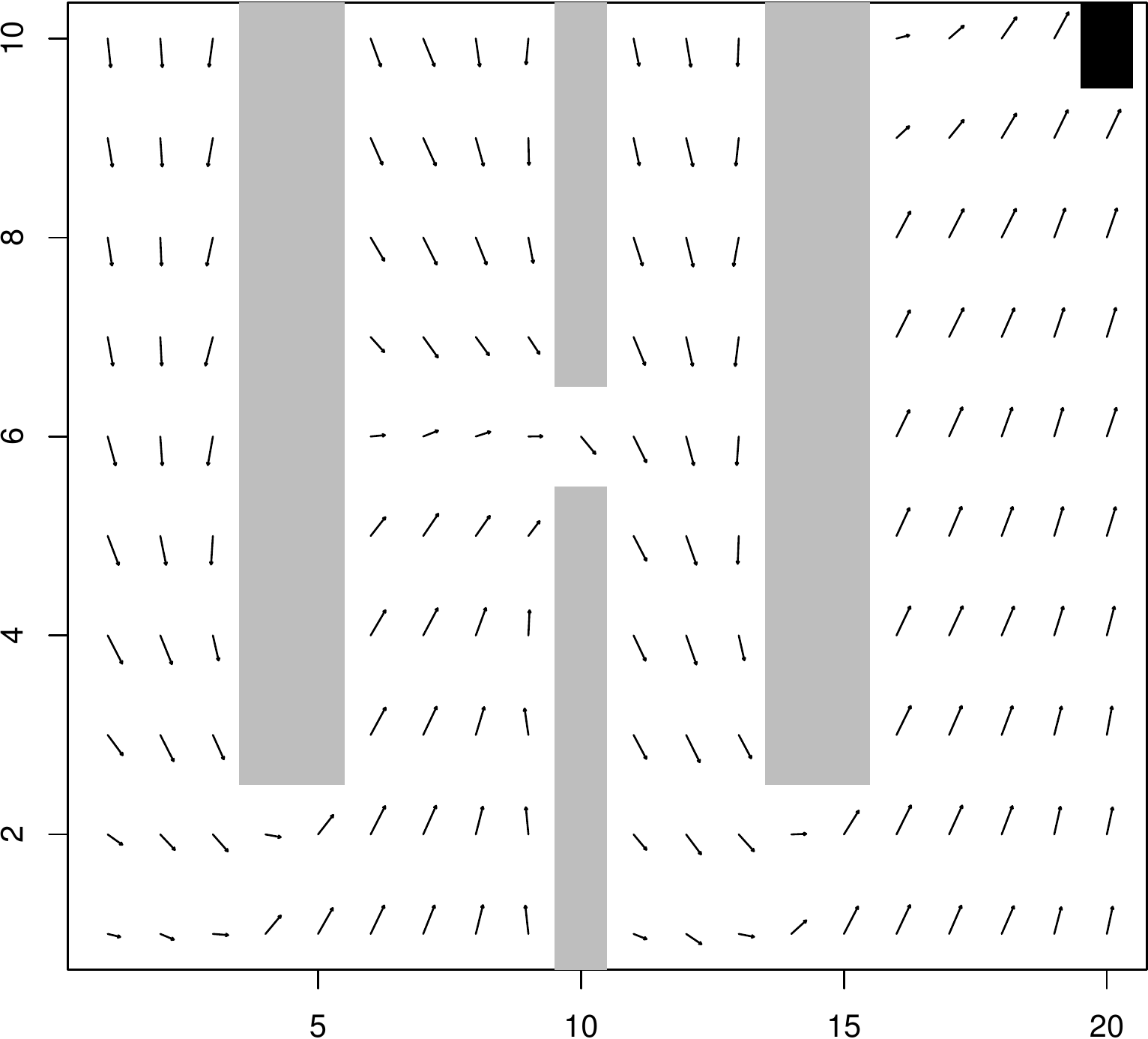}
\includegraphics[width=3in, height=1.5in]{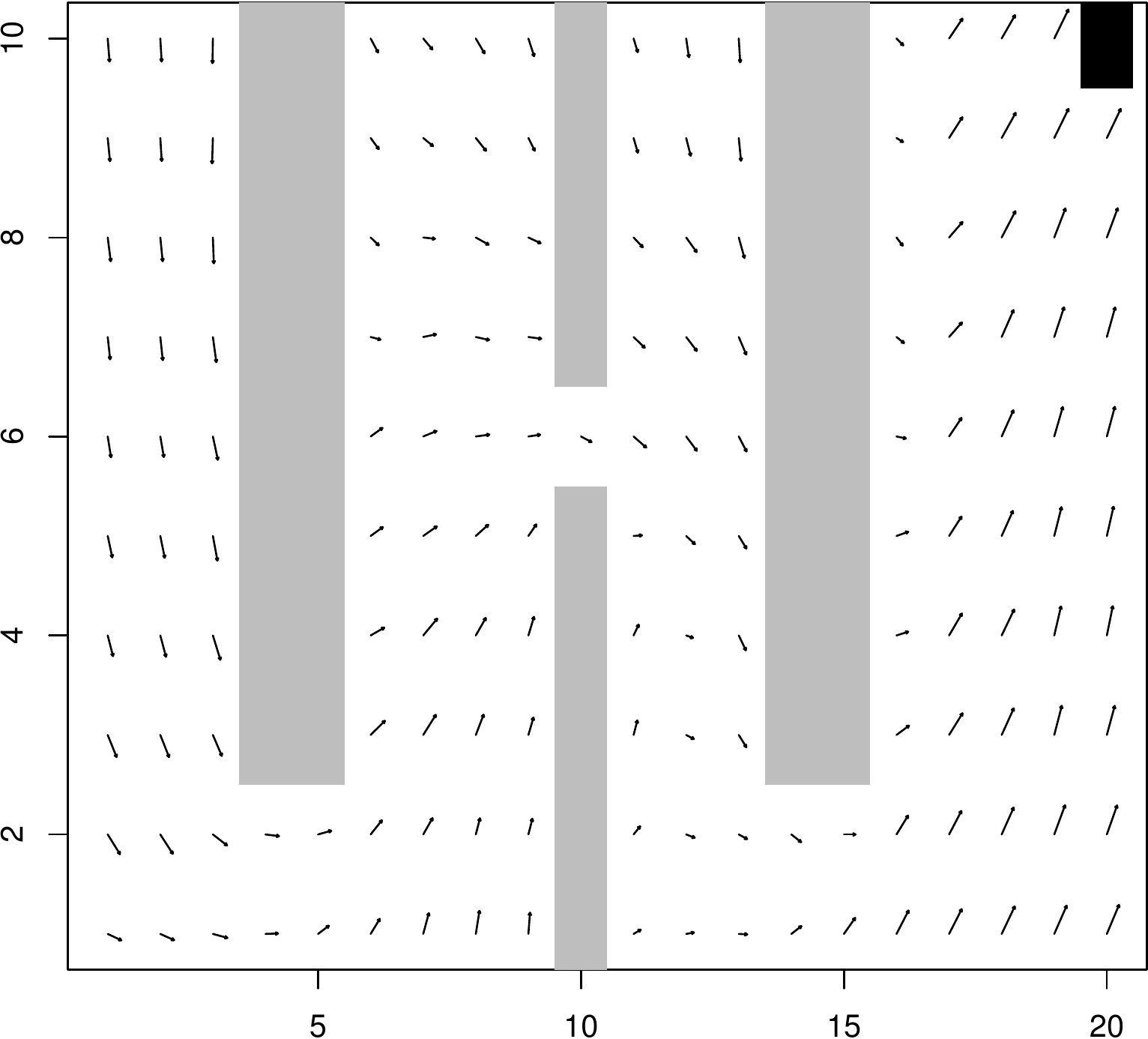}
\includegraphics[height=2.3in]{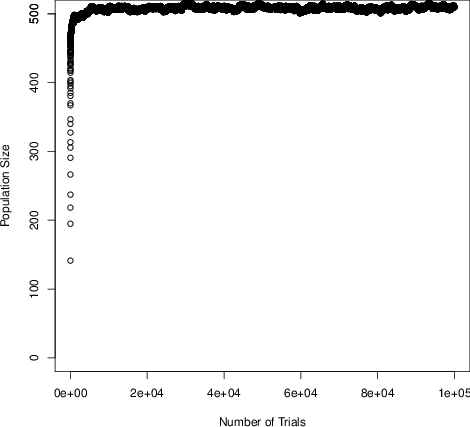}
\includegraphics[height=2.3in]{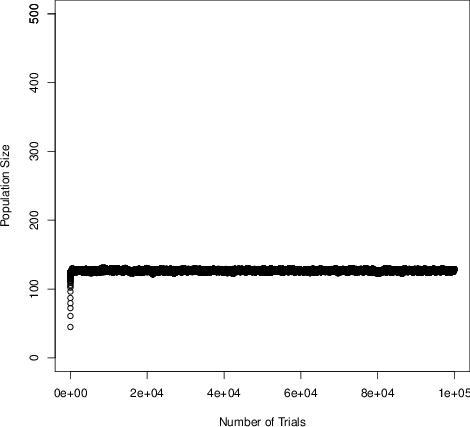}\\
\includegraphics[height=2.3in]{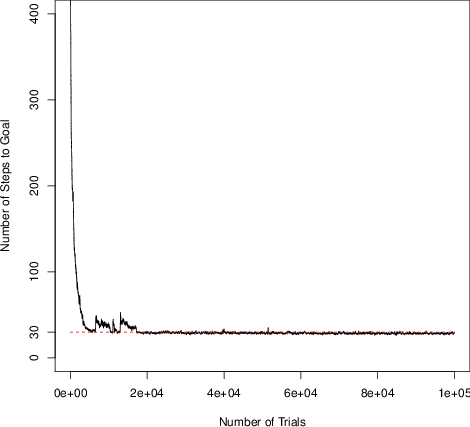}
\includegraphics[height=2.3in]{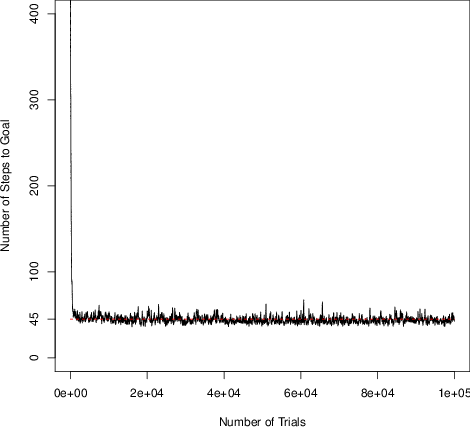}
\caption{SSOC2's behavior (upper row), population (middle row) and performance (lower row) throughout the trials. The left column shows the results for the default 10x10 sized population and the right show results for the smaller 5x5 SOM population.}
\label{big_maze1}
\end{figure*}

\begin{figure*}
\centering
\includegraphics[height=3in]{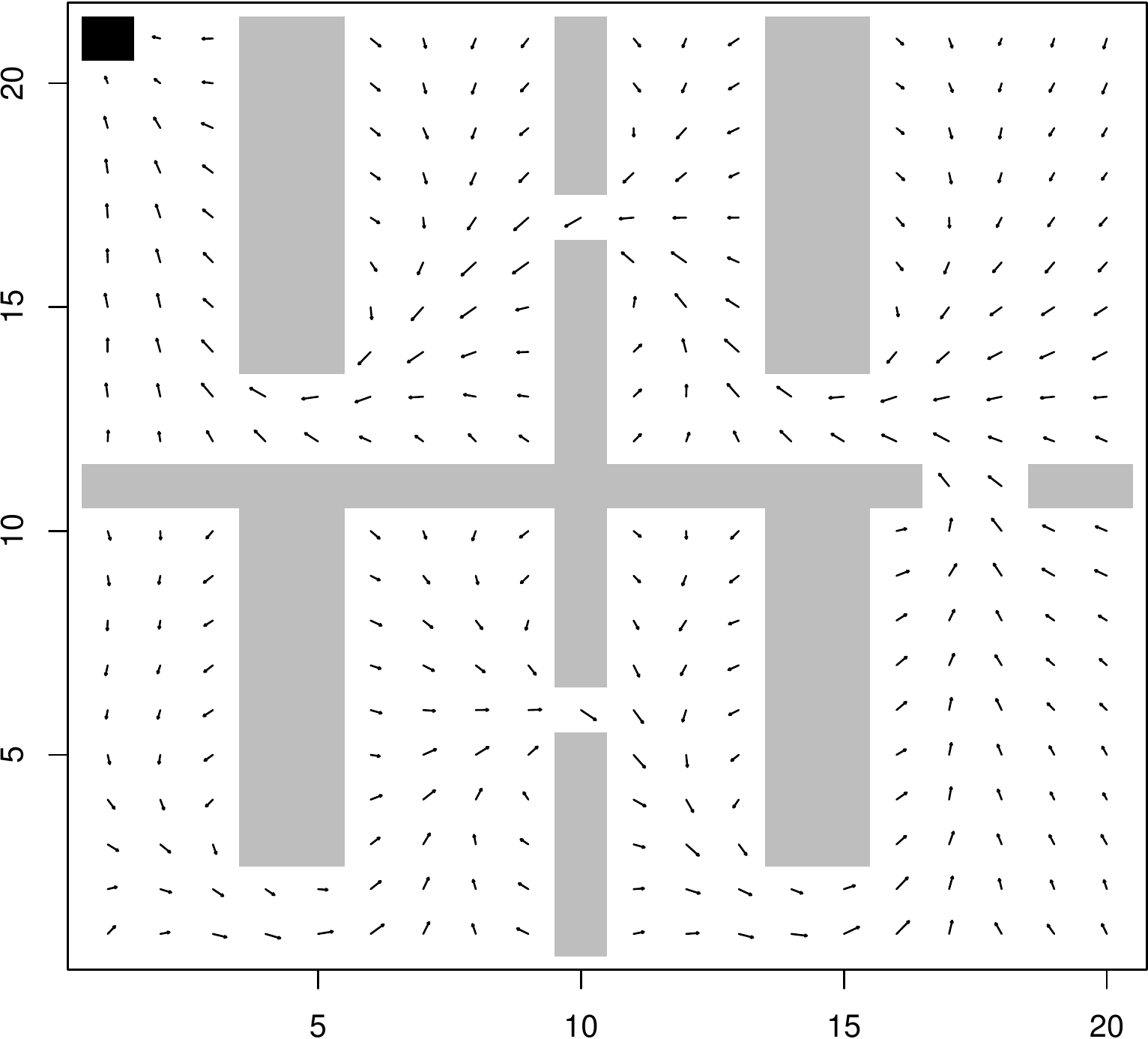}
\includegraphics[height=3in]{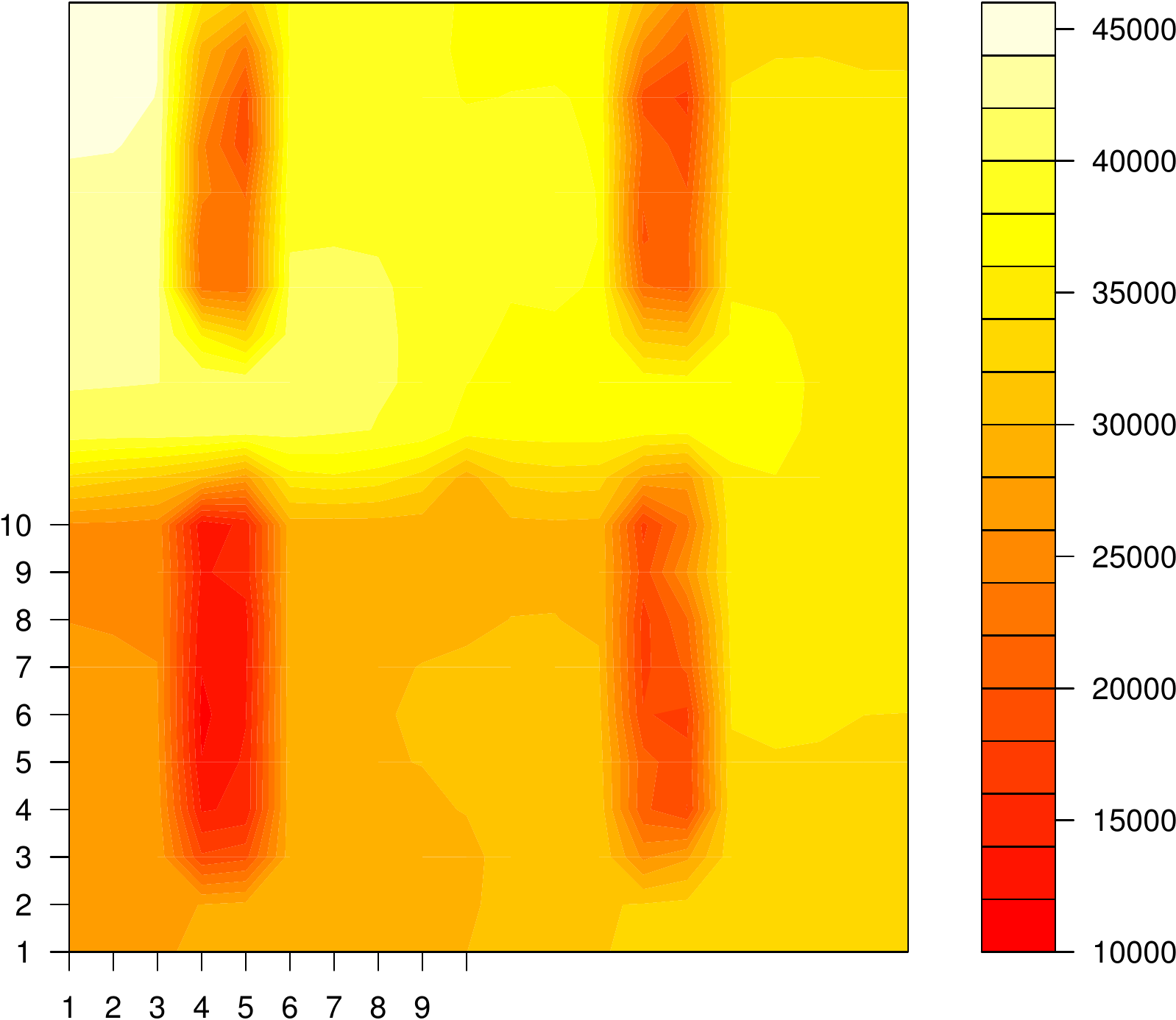}
\includegraphics[height=2.3in]{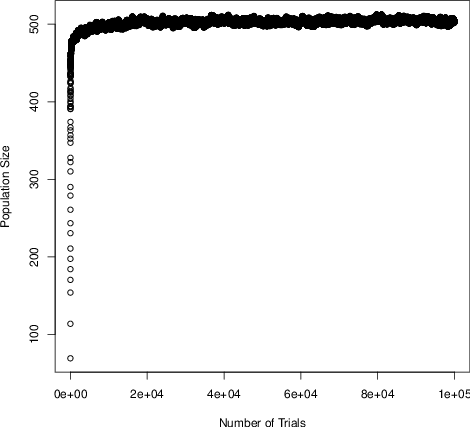}
\includegraphics[height=2.3in]{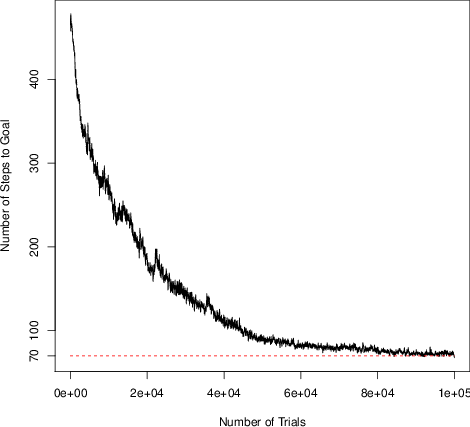}
\caption{SSOC2 applied to Maze $2$ with 20x20 SOM population and discount factor of $0.99$. 
The figure shows the behavior (upper left), fitness (upper right), population (lower left) and performance (lower right).}
\label{big_maze2}
\end{figure*}

\begin{figure*}
\centering
\includegraphics[height=3in]{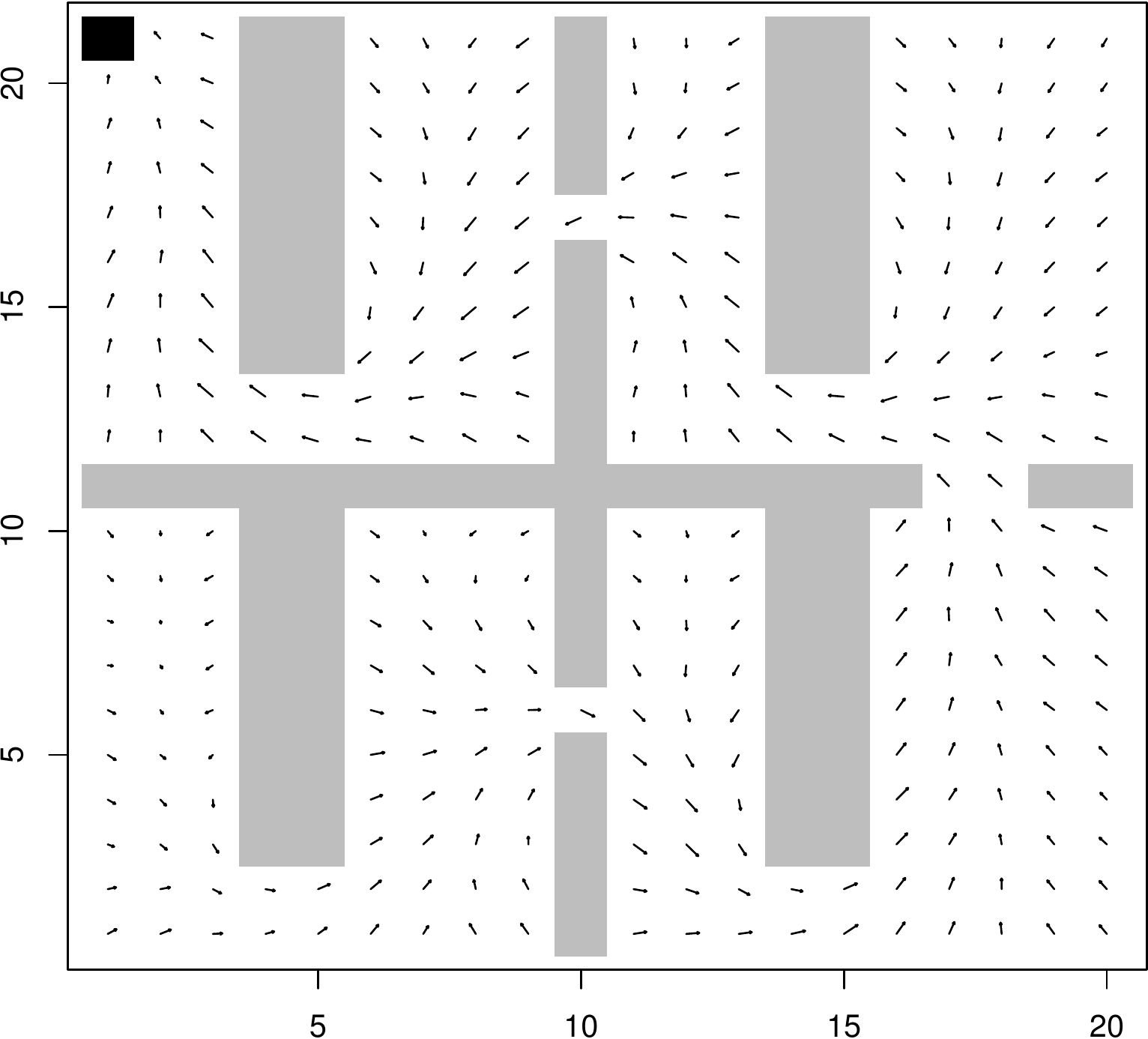}
\includegraphics[height=3in]{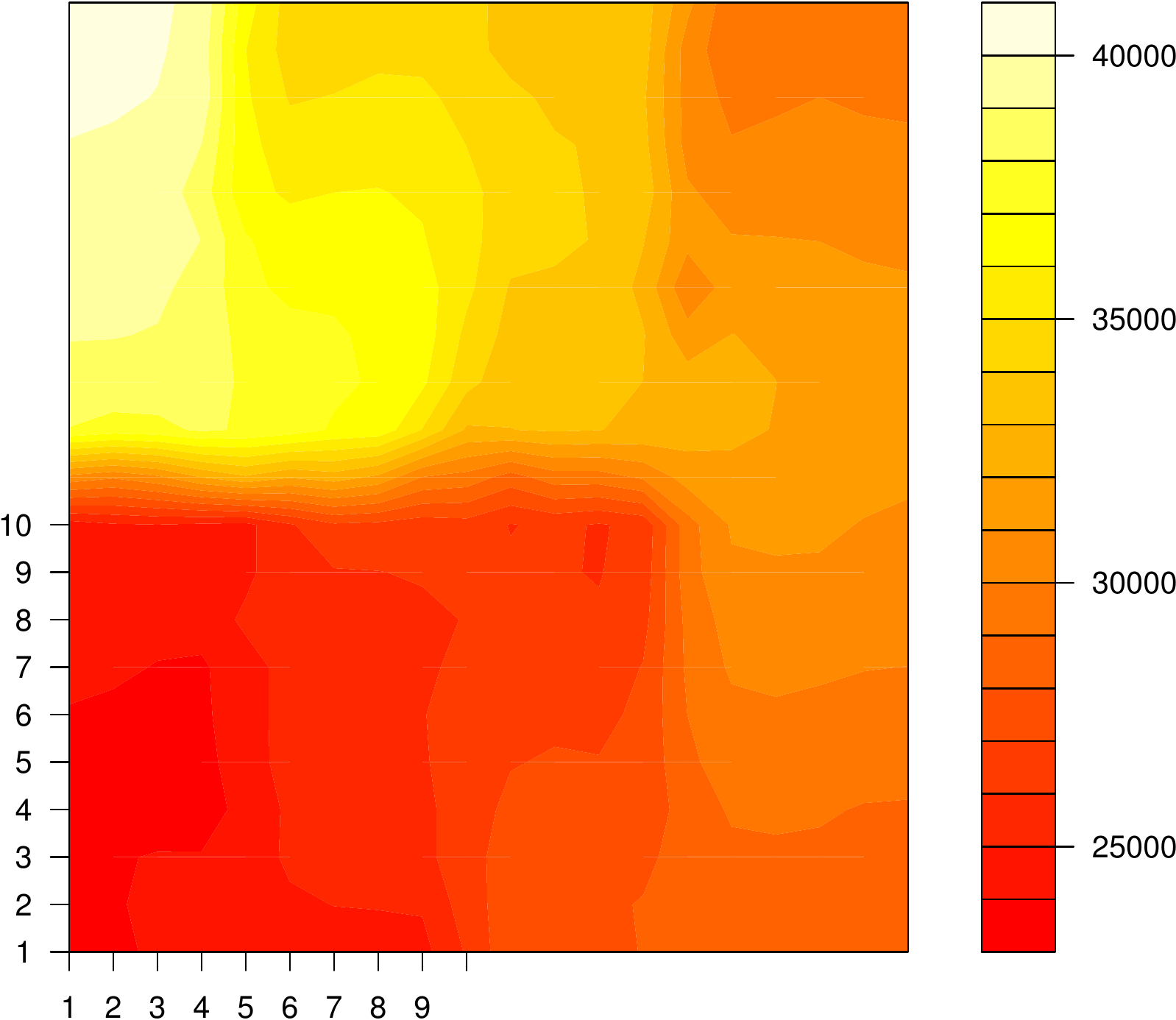}
\includegraphics[height=2.3in]{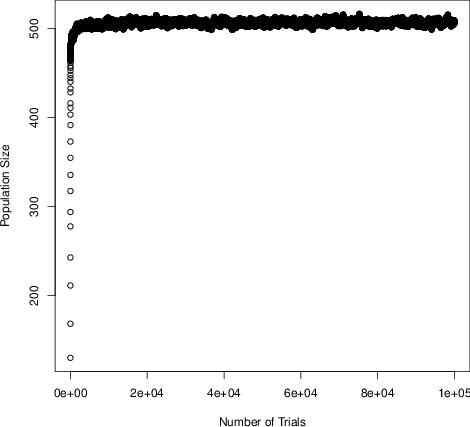}
\includegraphics[height=2.3in]{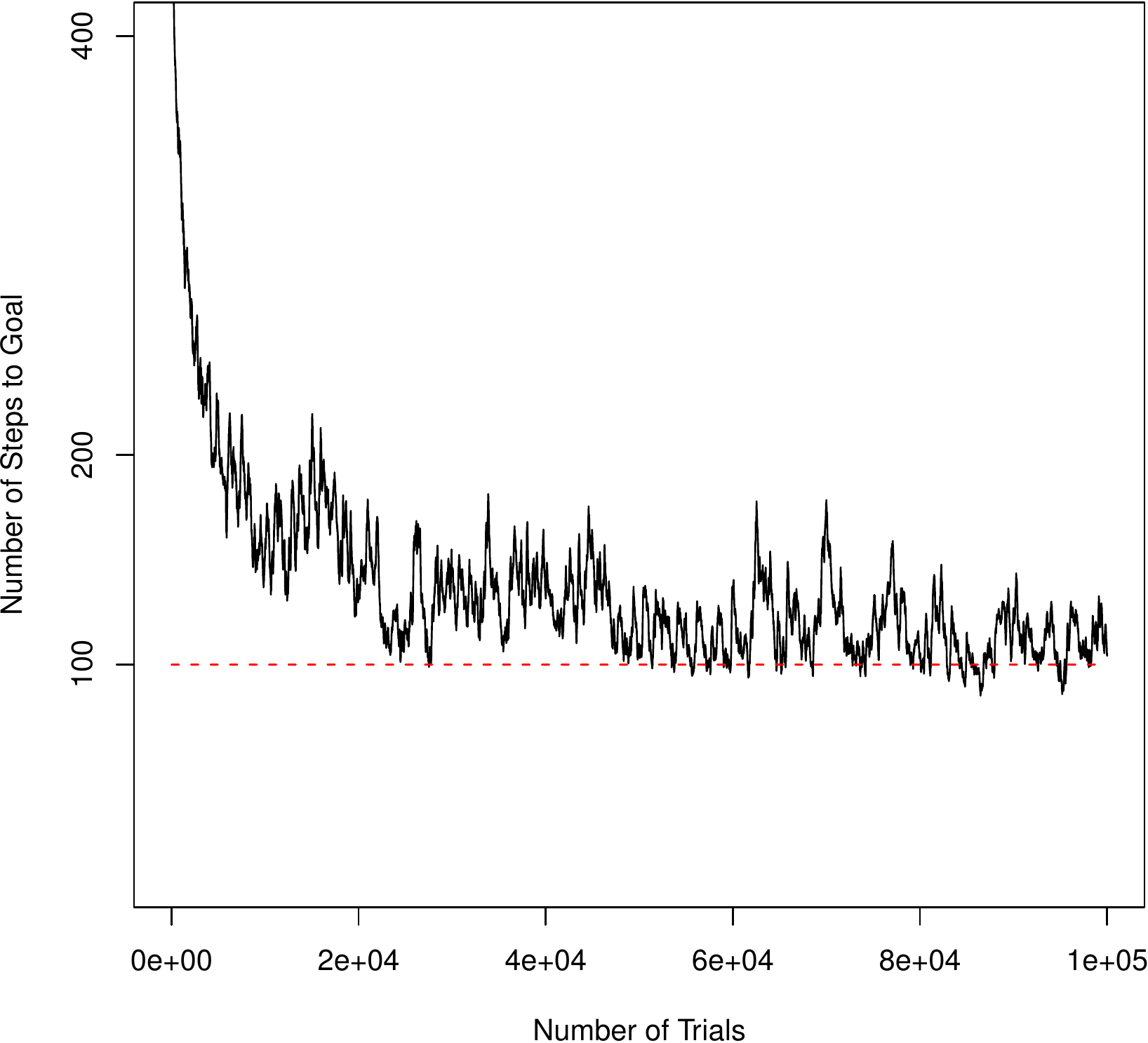}
\caption{Smaller population results. SSOC2 applied to Maze $2$ with a discount factor of $0.99$ (remaining parameters are the default, i.e., 10x10 SOM Population, etc).
The figure shows the behavior (upper left), fitness (upper right), population (lower left) and performance (lower right).}
\label{big_maze2_less_pop}
\end{figure*}

Figure~\ref{big_maze1} shows the behavior, population and performance of SSOC2 over Maze $1$ with both the default 10x10 SOM population and with a smaller 5x5 SOM population.
It can be seen that even with an insufficient population the algorithm achieves a solution capable of near optimal results.
The same can be said for the results over Maze~$2$ in Figures~\ref{big_maze2} and~\ref{big_maze2_less_pop}.
Despite the size of the maze, the quality of the result does not decrease much with smaller population sizes. 
In fact, the behavior stays more or less the same with the performance being a little worse. 
Both algorithms use a discount factor of $0.99$ to enable a good spread of fitness.

SSOC2's sensitivity to the size of population is existent but it is, surprisingly, not big even with such simple classifiers.
The sensitivity happens due to two limitations:
\begin{enumerate}
	\item The SOM's current incapability of growing to couple with the difficulty of the problem in question.
	\item SSOC2's classifiers are as simple as possible. 
		Therefore, niching's granularity needs to increase for the method to cope with different actions. 
		In fact, it is just evolving piecewise constant approximations (neural XCSF from \cite{howard2009towards} used piecewise nonlinear approximations to solve continuous input-action multi-step problems). 
\end{enumerate}
Therefore, with adaptable niching's granularity (growing and shrinking SOM population) as well as more complex classifiers (complex models) the sensitivity should disappear completely.

\subsection{Noisy mazes}

Real world problems always have some degree of noise involved.
In some of them, the noise present can not be disregarded, affecting some algorithms undesirably. 
To simulate mazes where noisy can not be disregarded, a $\pm5\%$ noisy variation is added to the observed variables.

\begin{figure}
\centering
\includegraphics[width=3in, height=1.5in]{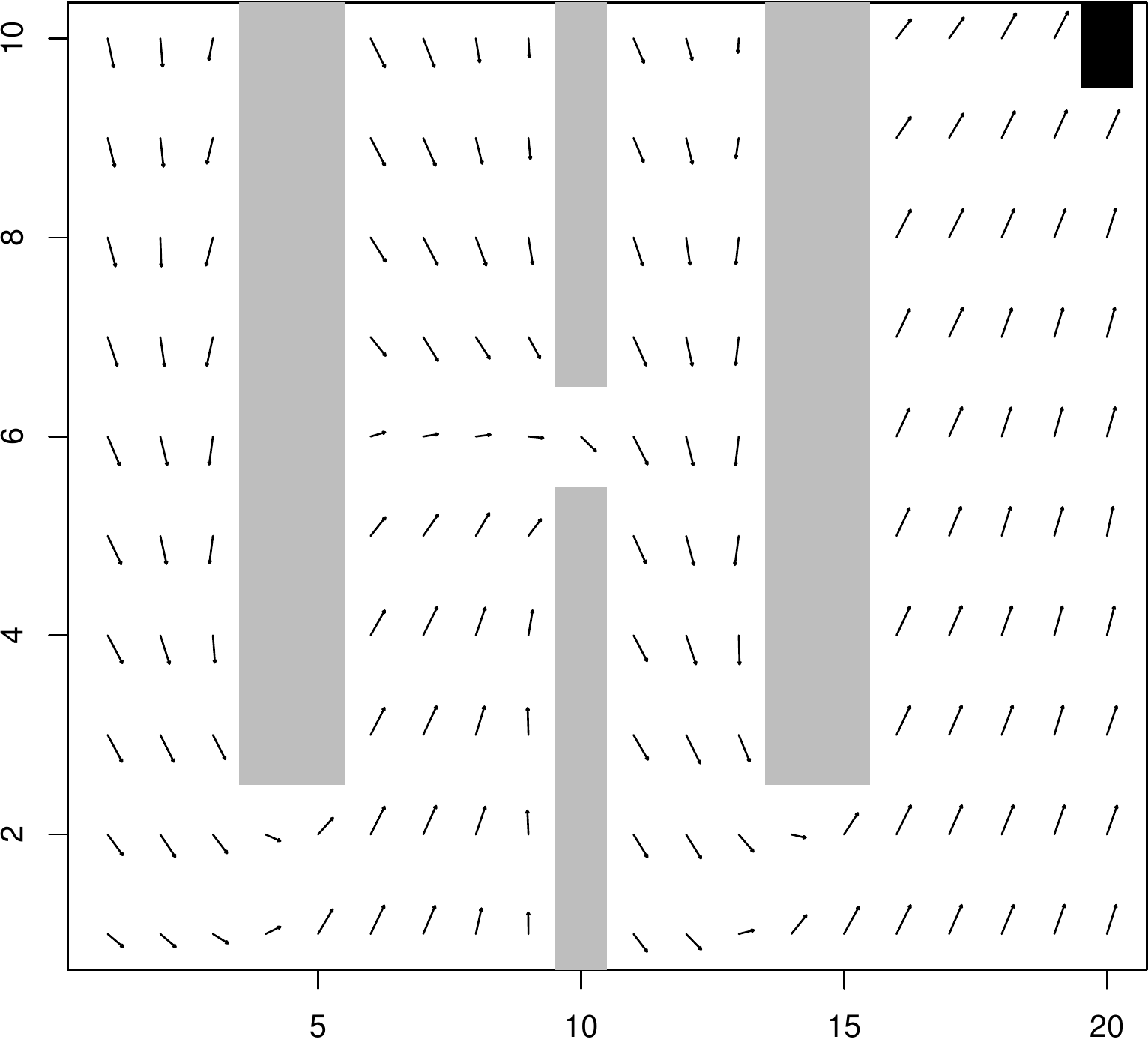}
\includegraphics[height=2.3in]{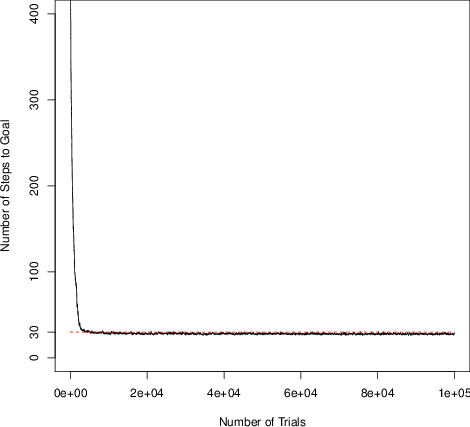}
\caption{SSOC2's behavior and performance for Maze 1 with noise.}
\label{noisy_maze1}
\end{figure}

\begin{figure}
\centering
\includegraphics[width=3in, height=3in]{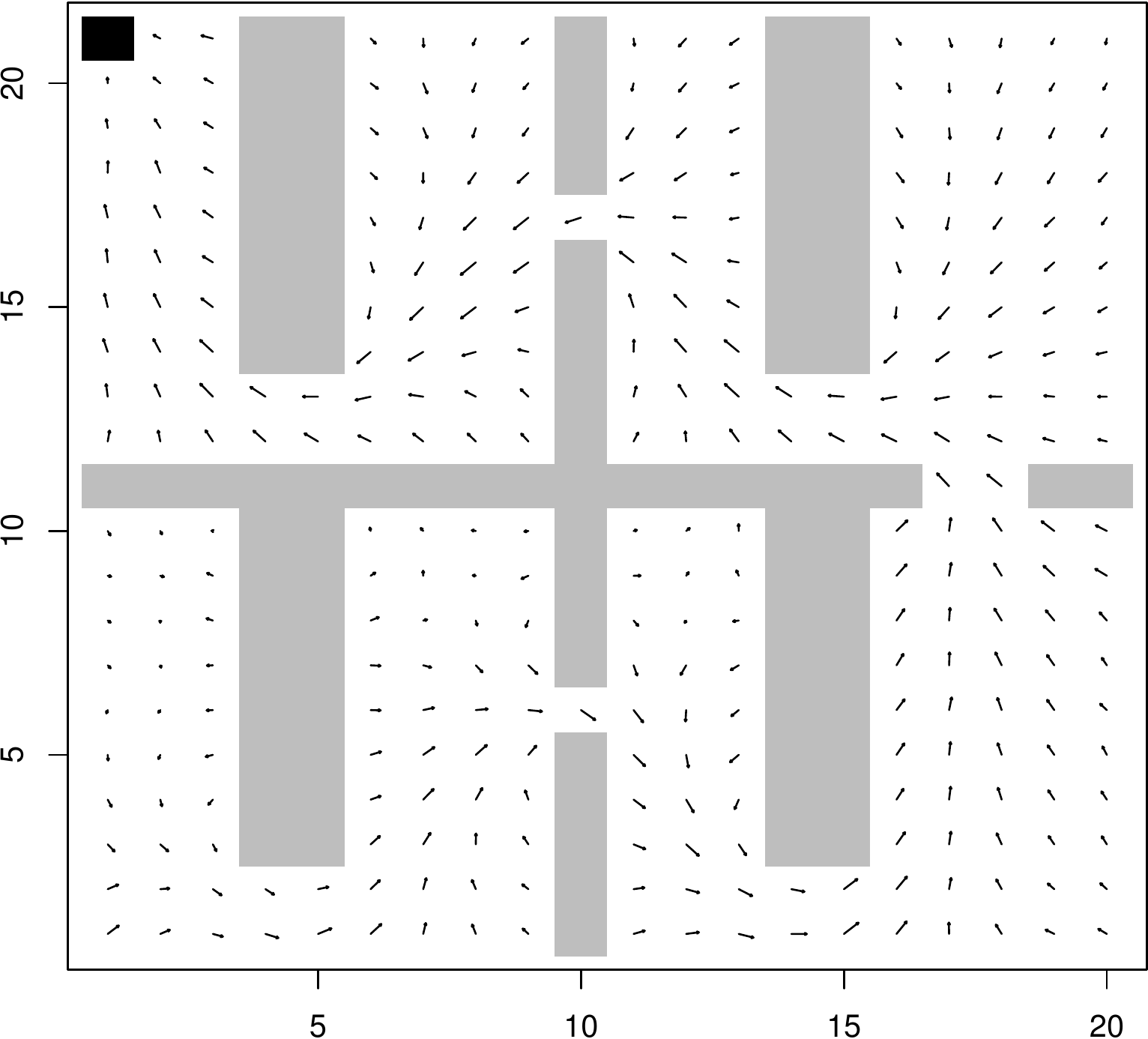}
\includegraphics[height=3in]{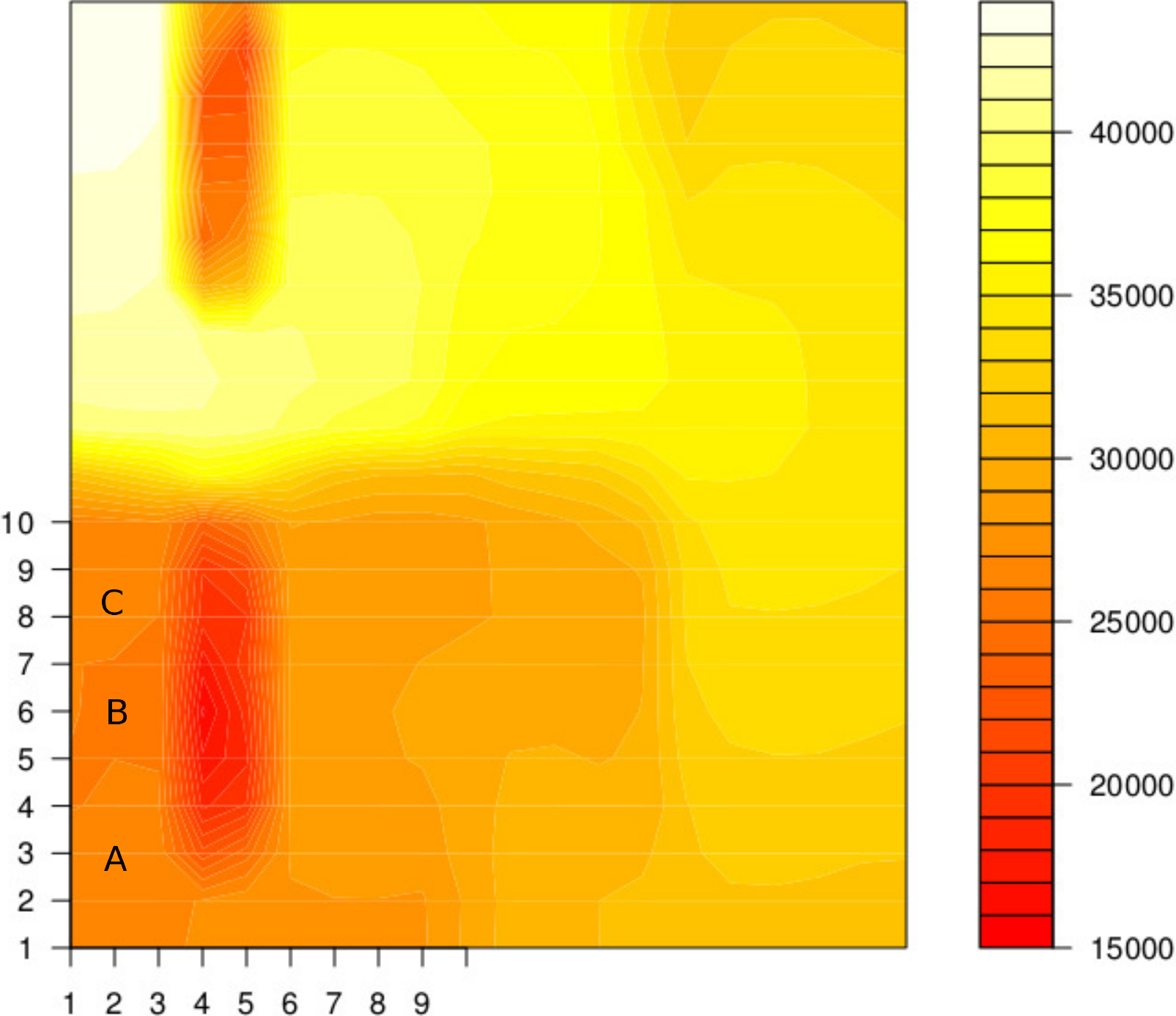}
\includegraphics[height=2.3in]{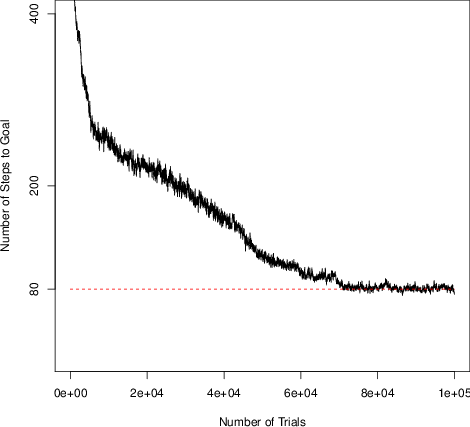}
\caption{SSOC2's behavior, fitness and performance for Maze 2 with noise. SOM population's size is 20x20 and the discount factor is set to $0.99$.}
\label{noisy_maze2}
\end{figure}

The result is shown for both Mazes 1 and 2 on respectively Figures~\ref{noisy_maze1} and~\ref{noisy_maze2}.
For Maze $1$, Figure~\ref{noisy_maze1} shows very good results with no visible issues. 
In fact, the performance is better than the results without noise (see Figure~\ref{big_maze1}), i.e., it has less oscillations while converging to the same value.
This happens because without noise the algorithm may get stuck, repeating for a long time the same input and consequently building a poor SOM model (the weights of SOM's cells will get near the repeated input). 
With noise the input received by the algorithm will rarely be the same.
Actually, the algorithm may even found itself fewer times stuck, since the noise is constantly changing the input giving chance to other SOM cells to activate and therefore the possibility of different actions to arise.
Maze $2$ shows a slightly worse performance and behavior.
This happens because aliasing states with very different fitness appeared.
To explain this phenomenon, first take a look at the fitness in part A, B and C.
Notice that parts A and C have greater values than part B.
This is only possible because part C is getting a bit of fitness from the high fitness northwest portion of the map just above it. 
Due to the noise effects, part C may receive inputs from the northwest portion of the map, allowing it to receive a greater reward than it should.
This influence of a high fitness aliasing state cause the unwise behavior at the frontiers.
However, when the initial state is not in this problematic regions, the result is as good as without noise.

\begin{figure}
\centering
\includegraphics[height=2.5in]{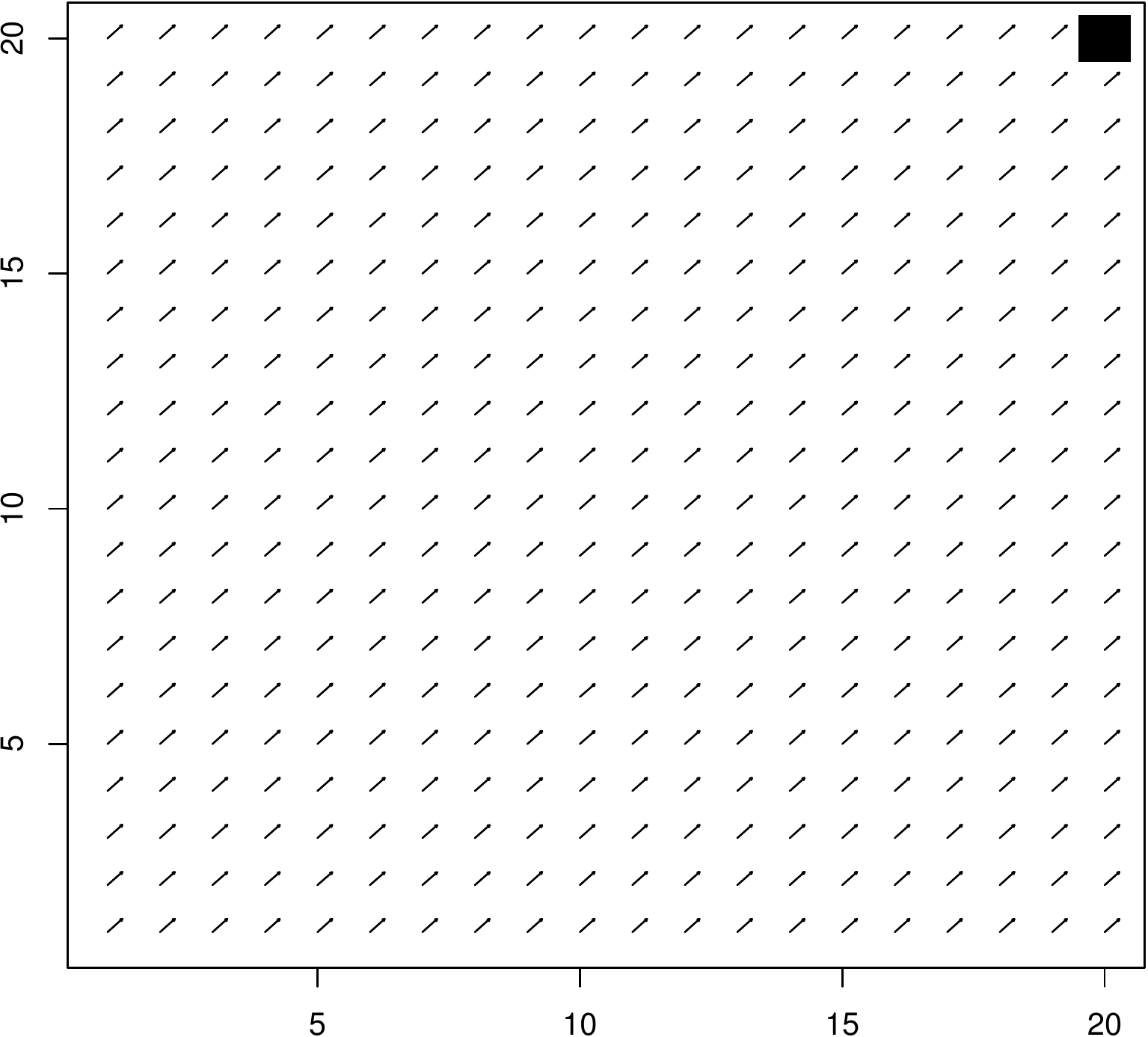}
\caption{Empty room maze with noise solved by SSOC2 with a SOM population's size of 5x5.}
\label{noisy_maze0}
\end{figure}

To give an idea of SSOC2's performance in relation to the evolutionary machine learning literature, Figure~\ref{noisy_maze0} shows the resulting behavior for the empty room with noise of \cite{howard2009towards}.
Both SSOC2 and the neural XCSF converge to the optimum behavior.
However, SSOC2 used a SOM population of size 5x5 which is equivalent to a maximum of $175$ individuals in a panmictic population (each cell of the SOM population has $7$ individuals, see Table~\ref{para}), while XCSF took $16000$ individuals to solve it, i.e., SSCO2 used $91$ times less population.
Both SSCO2 and XCSF were not optimized and these populations numbers are not the minimal population required to solve the maze. 
However it is still a quantitative measure that gives an overall picture of the population requirements for both algorithms on multi-step RL problems.

\subsection{Changing Mazes}

Problems in the real world are never static.
To reflect this challenging characteristic, we consider a maze (see Figure~\ref{mazes2}) which changes from time to time.
In other words, the agent's adaptation ability is put to test.

\begin{figure}
\centering
\includegraphics[height=2in]{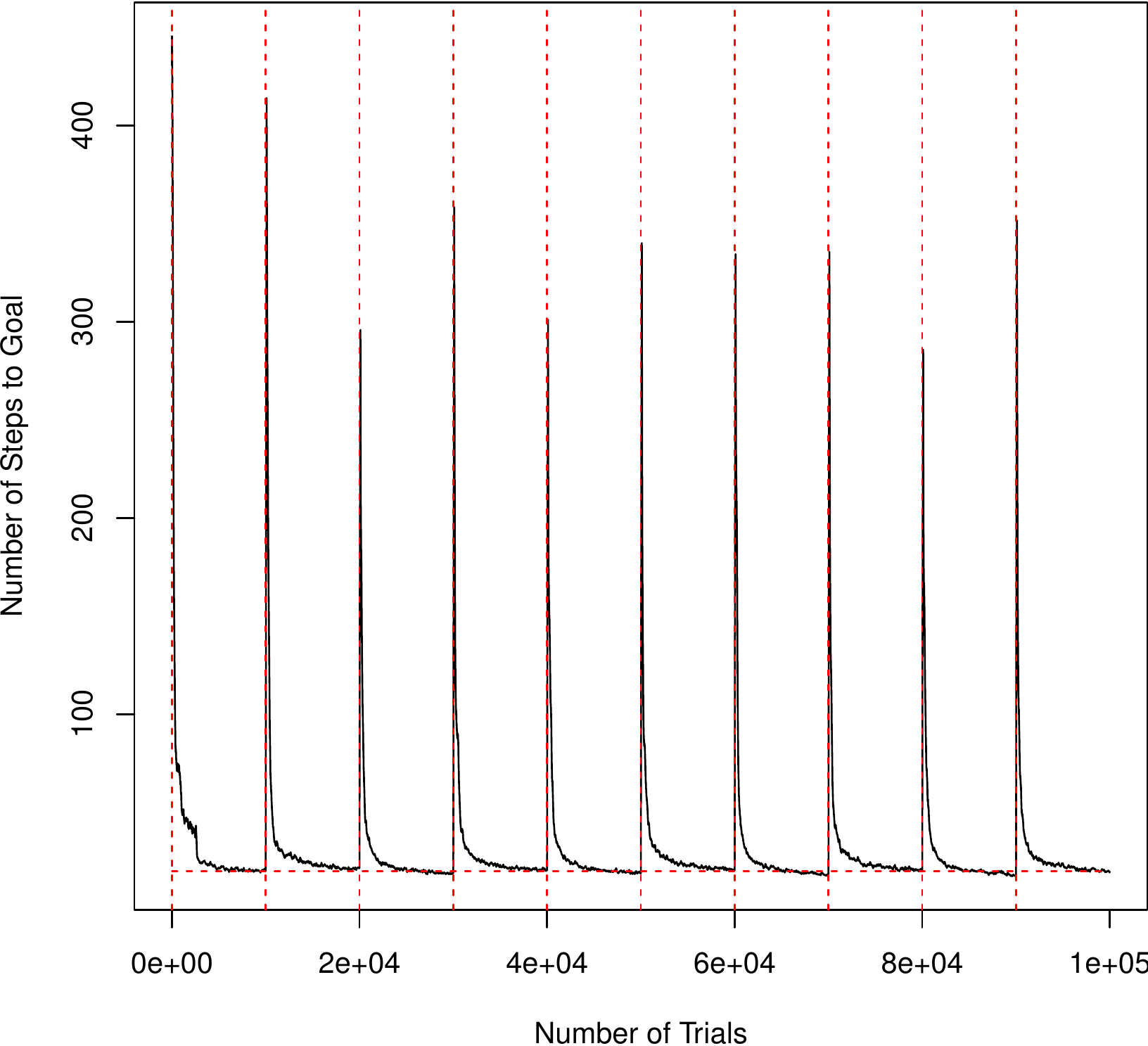}
\caption{Number of steps required by the agent to reach the objective in Maze~$3$. Vertical red dashed lines shows when the maze changed. A horizontal dashed line at $20$ is drawn for orientation purposes.}
\label{dynamic_result}
\end{figure}

Figure~\ref{dynamic_result} shows the number of steps required by the agent over time to solve Maze~$3$.
Notice that the first two peaks are bigger than subsequent ones.
This fact verifies the ability of the algorithm to reuse the knowledge when possible.
Moreover, the repetition of exponentially decreasing learning curves demonstrate the agent's ability to change its model to reflect the environment whenever the environment changes.
This verifies the agent's adaptation ability.

\subsection{Growing and compressing mazes}

A special case of dynamic mazes are mazes which does not only change its internal structure but also size.
That is the case for Maze~$4$'s growing maze (see Figure~\ref{mazes3}).

\begin{figure}
\centering
\includegraphics[height=2in]{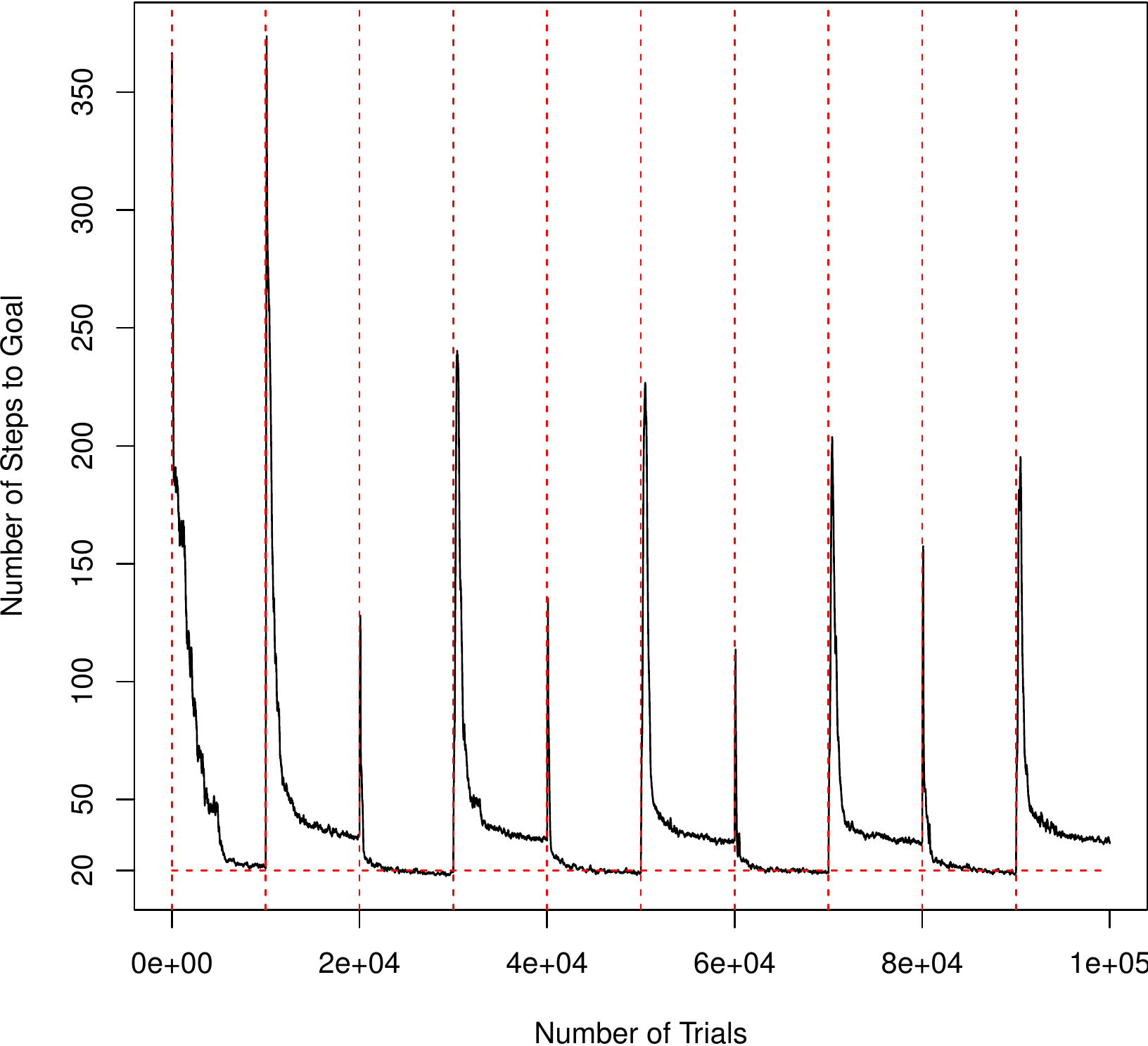}
\caption{SSOC2 performance on Maze~$4$. Vertical red dashed lines show when the maze changed. A horizontal dashed line at $20$ is drawn for orientation purposes.}
\label{grow_result}
\end{figure}

The results demonstrate that SSOC has a good capability of adaptation which is also unchanged throughout the experiment (see Figure~\ref{grow_result}).
Moreover, the decreasing peaks shows the reuse of knowledge from the previous problem. 
The system seems to arrive at a stage where minimal steps are required to adapt, i.e., minimal adaptation's time.

\section{SSOC versus SSOC2}

Here, SSOC2 and SSOC algorithms are compared.
In the SSOC algorithm, $0.1(0.999999)^{it}$ is used as a learning rate function (parameter absent in a parameterless SOM).

\begin{figure}
\centering
\includegraphics[height=2in]{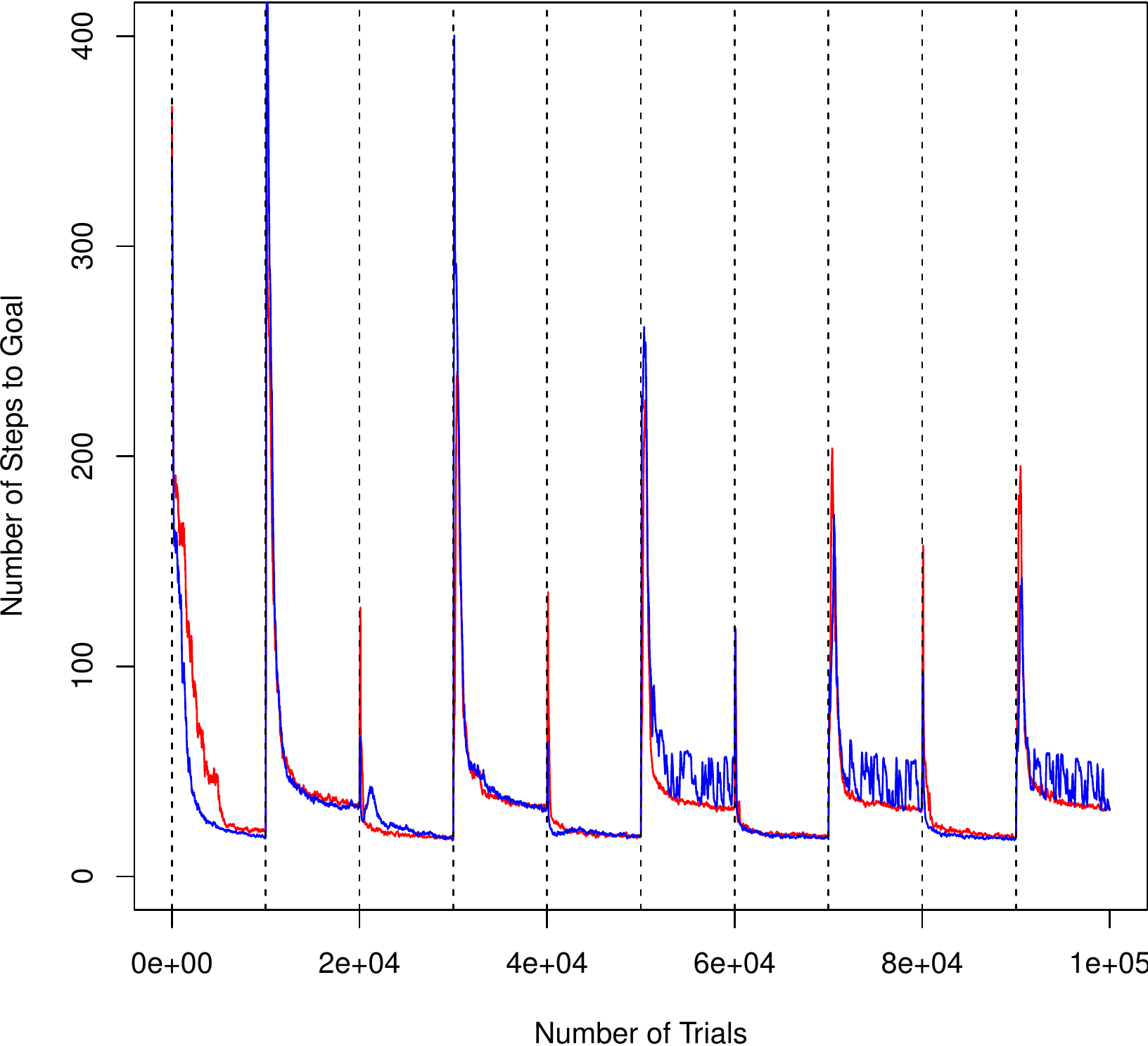}
\caption{Number of steps required by SSOC2 (red) and SSOC (blue) agents to reach the objective in Maze 4. Vertical dashed lines show when the maze changed.}
\label{grow_comp}
\end{figure}

\begin{figure}
\centering
\includegraphics[height=2.3in]{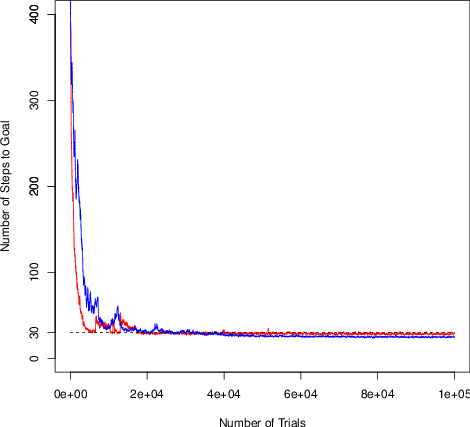}
\caption{Results on Maze 1 for SSOC2 (red) and SSOC (blue).}
\label{maze1_comp}
\end{figure}

\begin{figure}
\centering
\includegraphics[height=2.3in]{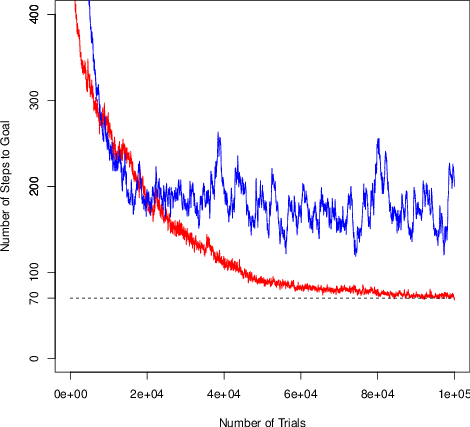}
\caption{Results on Maze2 for SSOC2 (red) and SSOC (blue).}
\label{maze2_comp}
\end{figure}

\begin{figure}
\centering
\includegraphics[height=2in]{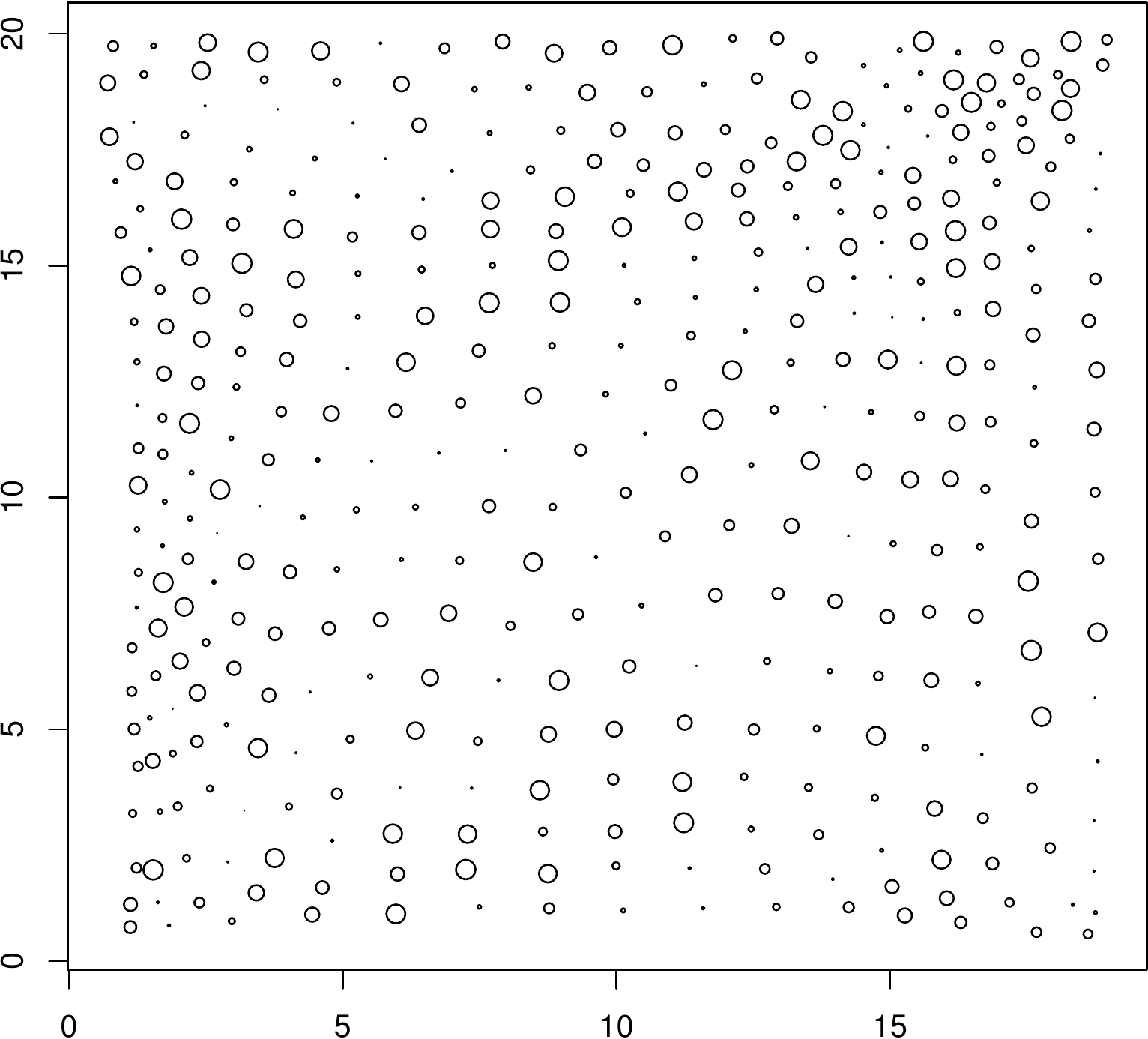}
\includegraphics[height=2in]{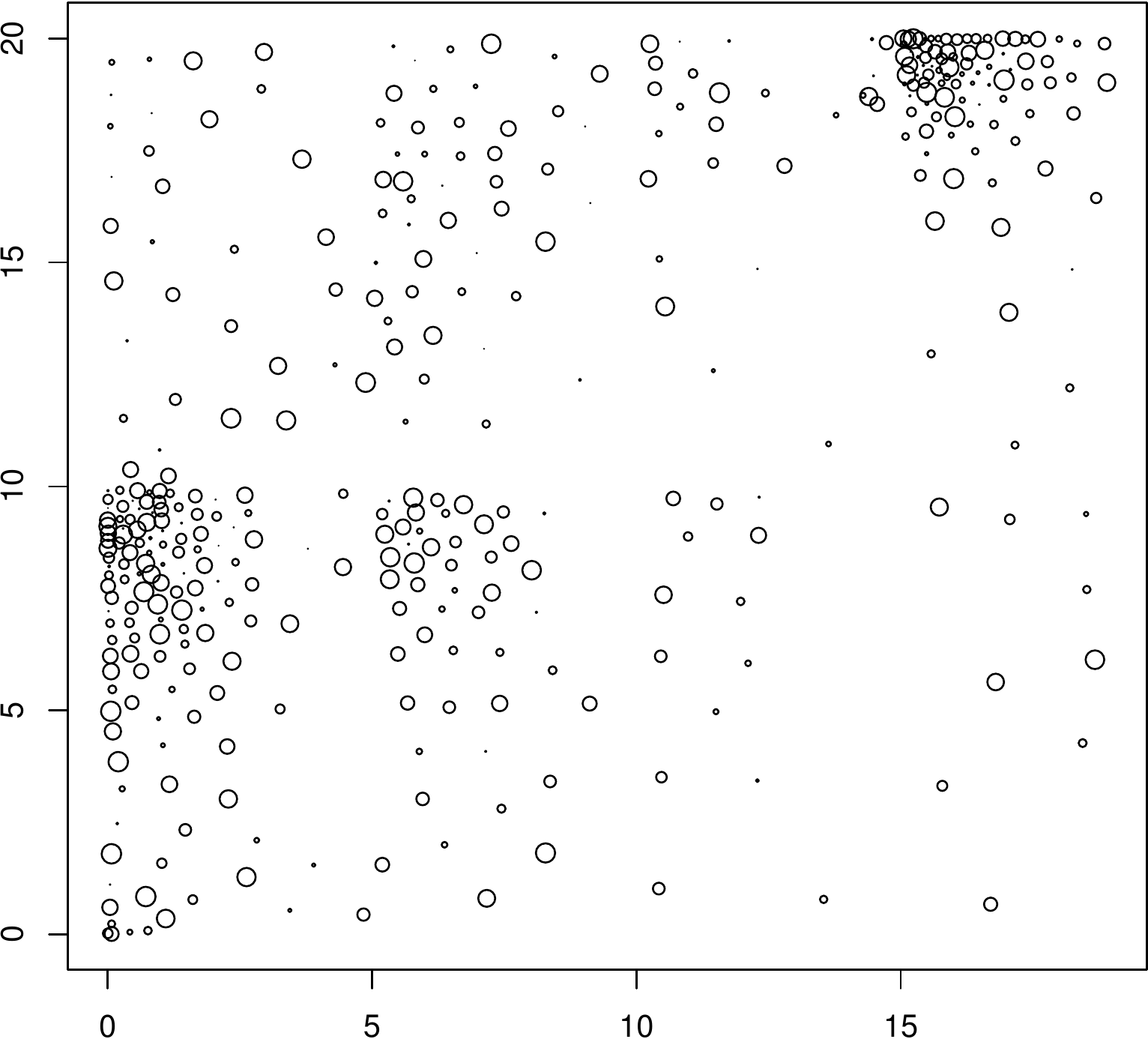}
\caption{Two samples of the final distribution of SOM's cells for SSOC2 (above) and SSOC (below) for Maze~$2$. The size of the points is directly proportional to their experience (number of times they were selected to act).}
\label{som_map}
\end{figure}

The main difference between SSOC and SSOC2 are the incapability of changing the model when an early distribution was biased (or when the problem has just changed).
Therefore, the results from Figure~\ref{grow_comp} are expected (SSOC was not able to continuously adapt as the learning rate gets smaller and can not increase with time).
Figures~\ref{maze1_comp} and~\ref{maze2_comp} shows respectively a smaller problem (solved by both algorithms) and a bigger problem where SSOC had difficulties.
Similar to problems that change with time, big mazes may cause the distribution of input from a series of trials to differ strongly. 
To make the differences explicit, Figure~\ref{som_map} shows a comparison between the resulting SOM's structure from both SSOC and SSOC2.
It is clear that SSOC focuses on the frequency of input while SSOC2 focuses on the novelty of the input \footnote{recall that the error between the SOM's cell and the input, which is used to determine the rate of learning, is an approximation to uniqueness, i.e., a measure of novelty \cite{ICDL12-hester}, \cite{reehuis2013novelty}}.
That explains the poorer coverage of SSOC. 
Having said that, a poor coverage does not mean a worse result. 
It depends on the system. 
For example, if the high frequency of inputs coincides with the most difficult inputs, the poor coverage would be a very useful mapping.

\section{Mixed Genetic Operator}

SOM's population is naturally a structure.
However, until now the evolutionary algorithm did not take any benefit from this aspect.
Here we show that by using the topological preserving properties of SOM it is possible to mix global and local solutions to obtain relevant improvements.

\begin{figure}
\centering
\includegraphics[height=2in]{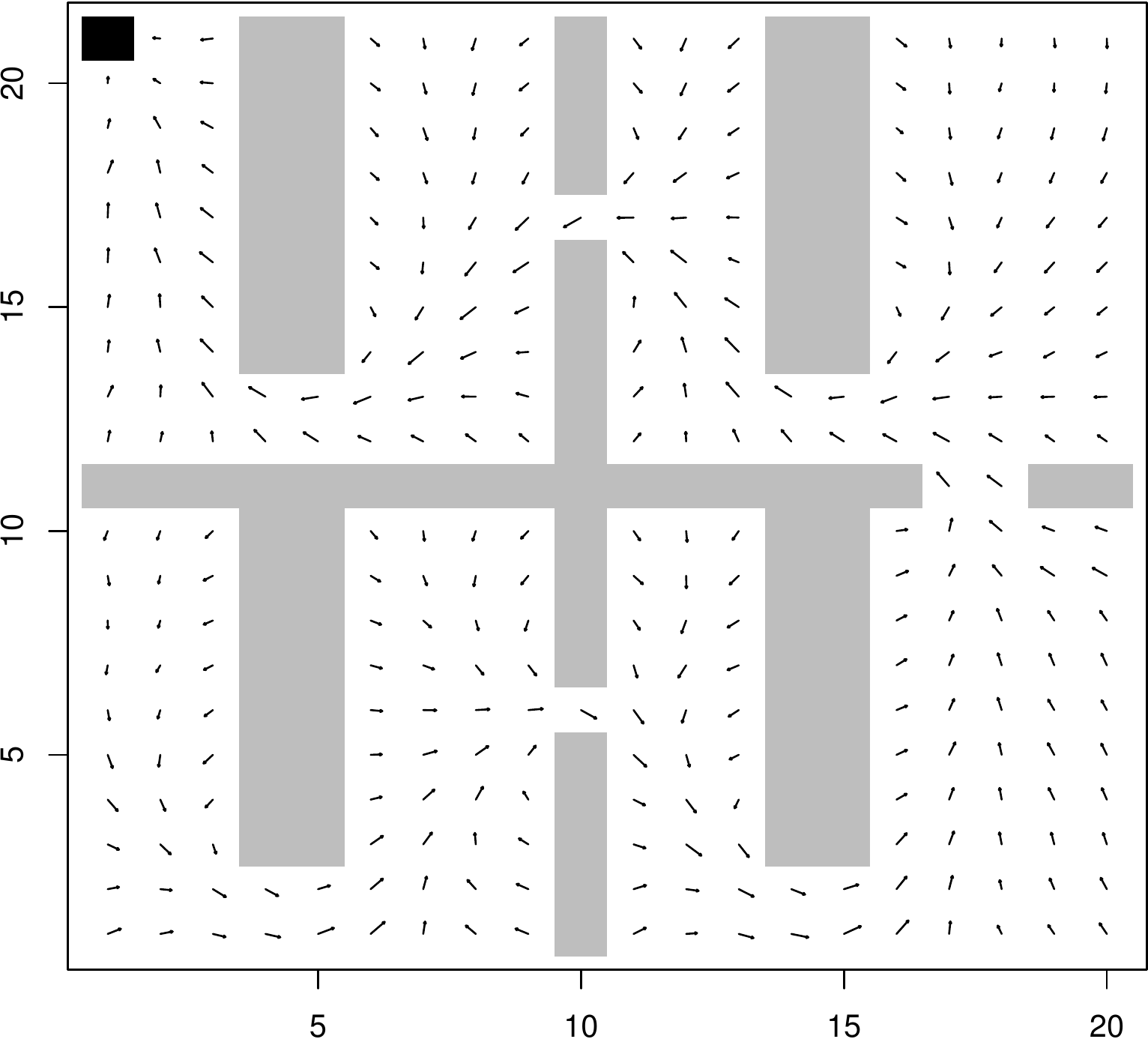}
\includegraphics[height=2in]{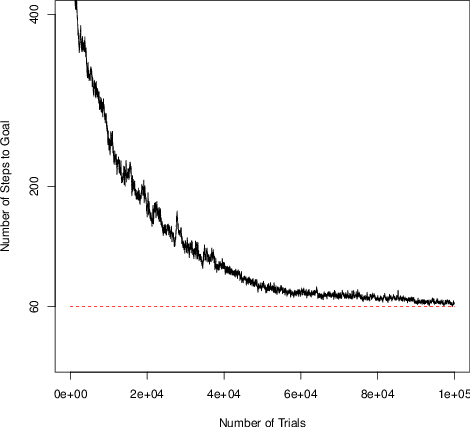}
\caption{SSOC2 with mixed genetic operator applied to Maze~$2$.}
\label{mixed_operator}
\end{figure}

The mixed genetic operator used is, as before, based on the differential evolution operator.
But instead of choosing three random global solutions, each of the three solutions chosen come with equal probability from either a random adjacent cell (local) or a random solution (global).
Results are shown on Figure~\ref{mixed_operator}.
The number of steps necessary to reach the goal is on average approximately $60$ after the algorithm has converged.
Recall that using the previous (only global) genetic operator the average reached approximately $70$ steps (see Figures~\ref{big_maze2}).
This result shows that evolutionary algorithms can explore the structure of the SOM with genetic operators, ending up also working on the inherent structure of the problem.
In other words, it is possible to efficiently reuse similar partial solutions (adjacent solutions) to solve similar partial problems (adjacent cells).

\section{Justification}

SSOC2 has a simple model.
Nonetheless, it is still possible to achieve good results in complex problems with it.
One way to think about it (as described in Section~\ref{intro_sec}) is to see that niched fitness is necessary to avoid over-generalized solutions, deriving the rest of the reasoning from it.
However, one might also think that at the heart of any problem, there is the necessity to divide and decouple as much as there is the necessity to solve the parts.
In other words, divide and conquer\footnote{The name of a widely known algorithm but most importantly a line of thought of how to solve problems in general} is essential.
The hurdle is that it was always a multi-objective problem, treating it as a single-objective problem will consequently provoke either division or conquer strategies to prevail over another, while we may want both to coexist\footnote{over-generalizing solutions derives exactly from the prevalence of conquer strategies over division strategies}.
By separating both niching and fitness pressures into a cooperating system, SOC managed to overcome this problem.
In this sense, niched fitness becomes a consequence.

\subsection{The Mistake in the Metaphor}

LCS are based on an amazing metaphor where individuals should behave as well as choose their own niche.
There is no mistake in this metaphor.
The problem comes when fitness is defined as an exaggeratedly simple single function per individual.
In this point, niches and other interesting metaphors are discarded because the system is using an absolute fitness, while the fitness should be relative to the niches not only numerically but in meaning (there is no meaning in combining a fitness from a niche with a fitness of another niche, they are variables from different attributes).
Monkeys may be amazingly adaptive in forests and many other environments, however, they are never able to interfere in any sense with the lantern-fish (fish of deep seawater).
That is the metaphor mistake that niched fitness attempts to solve.

\section{Conclusion}

This article evaluated deeply the benefits and challenges of self organized classifiers (or more generally structured evolutionary machine learning). 
The following are the main points:
\begin{itemize}
	\item SSOC2 - An improved version of SSOC (named SSOC2) that was created with the substitution of the SOM by the parameterless SOM. Differences in the final SOM population structured was shown as well as better results.
	\item State-of-art Results On Challenging Problems - Results were shown in a variety of continuous input-action multi-step problems. Problems with noisy and with dynamic environments were also considered. To the knowledge of the authors these problems are the most difficult continuous input-action multi-step problems faced by an evolutionary machine learning algorithm to date. Although SSOC2 is made of very simple classifiers (SSOC2 is capable of only piecewise constant approximations), near optimal results were presented. 
	\item Genetic Operator - Tests have shown the advantages of using the SOM population's topological information inside the evolutionary algorithm. By using global and local solutions inside the genetic operator, the average number of steps required to reach the objective reduced in approximately $10$ steps.
\end{itemize}

Thus, with the good results on very difficult problems it was possible to verify the strength of the approach even with a very simple internal model.
There are many possible branches of research that can derive from SOC. 
In fact, structured evolutionary machine learning algorithms were barely researched.
Numerous widely different algorithms are possible from the combination of parallel (structured) evolutionary algorithms, learning classifier systems and machine learning.
This is just the first step.



%
\bibliographystyle{abbrv}
\bibliography{sigproc}  
%
%
\end{document}